\documentclass{article}

\PassOptionsToPackage{numbers,compress}{natbib}

\usepackage[preprint]{neurips_2026}

\usepackage{multicol}
\usepackage{arydshln} 
\usepackage{graphicx} 
\usepackage{bm}
\usepackage{graphicx}  
\usepackage{subfig} 
\usepackage{authblk} 
\usepackage{wrapfig}
\usepackage{subfloat}
\usepackage{algorithm}
\usepackage{algorithmicx}
\usepackage{algpseudocode}
\usepackage{float}
\usepackage{multirow}
\usepackage{amsmath}
\usepackage{arydshln}
\usepackage{nicematrix}
\usepackage{caption}
\usepackage{setspace}
\usepackage{soul}
\usepackage{enumitem}
\usepackage{xcolor}

\usepackage[utf8]{inputenc} 
\usepackage[T1]{fontenc}    
\usepackage{hyperref}       
\usepackage{url}            
\usepackage{booktabs}       
\usepackage{amsfonts}       
\usepackage{nicefrac}       
\usepackage{microtype}      
\usepackage{xcolor}         

\title{Fused Bayesian Flow Networks for Dual-Target Molecular Design}

\newcommand*{\affaddr}[1]{#1} 
\newcommand*{\affmark}[1][*]{\textsuperscript{#1}}
\newcommand*{\email}[1]{\texttt{#1}}

\author{%
\textbf{Jingyuan Zhou}\affmark[1], 
~\textbf{Shikui Tu}
\affmark[1]\thanks{Correspondence authors are Shikui Tu and Lei Xu.}~, ~\textbf{Lei Xu}\affmark[1,2]$^*$\\

\affaddr{\affmark[1]School of Computer Science, Shanghai Jiao Tong University}\\
\affaddr{\affmark[2]Guangdong Laboratory of Artificial Intelligence and Digital Economy (SZ)}\\
\email{\{zjoyuan0930, tushikui, leixu\}@sjtu.edu.cn}\\
}

\begin{document}

\maketitle

\begin{abstract}
Dual-target drug design aims to generate 3D molecules that can simultaneously interact with two target proteins, offering a promising route for discovering polypharmacological compounds against complex diseases. While recent generative models have shown encouraging performance in single-target drug design, existing dual-target approaches either focus on sequence generation or introduce an additional predictive drift term into the diffusion-based generative trajectory, which limits their ability to fully integrate feature information from both targets. We propose FusedBFN, a fused Bayesian flow network (BFN) for dual-target molecular design. FusedBFN formulates dual-target generation as distribution fusion in a unified continuous parameter space and employs a product-of-experts formulation to incorporate dual-target information throughout the generative process. To address the scarcity of dual-target structural data, we leverage a pretrained target-aware BFN model as the shared backbone. We further introduce a chemically aware prior-based alignment method and a prior-free pocket alignment strategy to construct aligned dual-target contexts. Extensive experiments demonstrate that FusedBFN generates molecules with strong binding affinity toward dual targets while maintaining favorable molecular properties.
\end{abstract}

\section{Introduction}

Structure-based drug design (SBDD) aims to generate small molecules that geometrically and chemically complement  the 3D structure of a target protein, thereby forming energetically favorable interactions \citep{klebe2025protein}. It represents a  rational and well-motivated approach for applying deep learning to drug discovery \citep{bai2024geometric}. In recent years, generative models have achieved promising results in this task; however, most existing methods focus on the single-target scenario, following a “one target, one drug” philosophy \citep{morphy2004magic, bolognesi2016multitarget, chaudhari2017computational}. Although this strategy has led to numerous therapeutic successes \citep{munson2024novo}, the complexity of biological networks \citep{gerstein2012architecture} implies that many diseases involve multiple factors and interacting pathways \citep{yuan2020ligbuilder}. Consequently, drugs acting on a single target are often inadequate to effectively treat complex diseases \citep{l2013polypharmacology, ramsay2018perspective, he2016combination}. Recently, there has been increasing interest in dual-target drug design \citep{srinivasan2024generation}, which seeks to develop a single ligand capable of simultaneously interacting with two distinct biological targets, following a “two targets, one drug” paradigm. Compared with single-target molecules, dual-target drugs can effectively enhance therapeutic efficacy, reduce the likelihood of resistance development, and simplify treatment regimens \citep{ye2023therapeutic, munson2024novo, francucci2025beyond}.

In contrast to the substantial body of work on 3D SBDD for single targets \citep{2021sbdd, 2022pocket2mol, targetdiff, guan2024decompdiff, qian2024kgdiff, gu2024aligning, huang2024protein, qu2024molcraft, zhou2025multi, zhou2025prior}, current research on dual-target drug design primarily focuses on sequence-based approaches without leveraging structural information from both targets. For example, Isigkeit et al. \citep{isigkeit2024automated}, Srinivasan et al. \citep{srinivasan2024generation} generate SMILES strings by fine-tuning chemical language models to explore the chemical space of dual-target compounds. In recent work, Zhou et al. \citep{zhou2024reprogramming} proposed a structure-based 3D framework for dual-target molecular design. Building upon a pretrained single-target diffusion model, their approach introduces an additional predictive drift term into the reverse generative trajectory to generate dual-target molecules. However, this drift-based strategy forcefully shifts the single-target distribution toward a dual-target objective, without fully integrating the distinct structural features from both binding sites. Furthermore, it aligns the two binding pockets by matching protein–ligand interaction priors (i.e., the ligands). During this alignment process, all ligand atoms are treated as equally important, thereby disregarding their underlying chemical semantics.

To tackle the aforementioned challenges, we propose FusedBFN, a \textbf{Fused} \textbf{B}ayesian \textbf{F}low \textbf{N}etwork framework for dual-target drug design. Motivated by the mechanism of Bayesian Flow Networks (BFNs) \citep{graves2023bayesian}, which updates the parameters of data distributions rather than the data itself via Bayesian inference, we formulate the task as fusing the distributions conditioned on two target protein contexts within a unified continuous parameter space. Specifically, we model the sender distribution under dual-target constraints using a product-of-experts (PoE) \citep{hinton2002training} formulation, which forms the starting point of the fusion process. The Bayesian updating process then propagates the fused target-context information into the parameter space, influencing the parameter updates and ultimately enabling the generation of molecules capable of interacting with both targets. However, due to the scarcity of structural data for dual-target–ligand complexes, directly training a dual-target generative model is impractical. To address this limitation, we leverage a pretrained target-aware BFN model \citep{qu2024molcraft} as the backbone, and extend the knowledge learned from single-target datasets to the dual-target setting through the fused Bayesian flow mechanism. When aligning the two binding pockets, we consider the chemical semantics of different atoms in the protein–ligand interaction prior and additionally propose a prior-free pocket alignment method to simplify the alignment process. Extensive experiments on the dual-target benchmark demonstrate that FusedBFN can generate molecules with high affinity to both targets while maintaining favorable molecular properties.

Our main contributions can be summarized as follows:
\begin{itemize}
    \item A novel distribution fusion framework for dual-target molecular design, which integrates distributions conditioned on two different targets in the parameter space to effectively fuse structural information from two binding pockets, thereby generating molecules capable of binding to both targets.
    \item To the best of our knowledge, this is the first formulation of a fused Bayesian flow network for dual-target generation, which directly leverages pretrained single-target models without requiring additional training or fine-tuning.
    \item A chemically aware dual-target alignment strategy that accounts for atomic chemical semantics when aligning protein–ligand interaction priors, together with an additional pocket alignment method that operates without ligand priors.
\end{itemize}

\section{Related Works}

\paragraph{Single- and Dual-Target Drug Design}

 Current SBDD methods are typically formulated for single-target scenarios. With advances in 3D and geometric modeling, numerous approaches have attempted to address this task directly in three-dimensional space. Ragoza et al. \citep{ragoza2022ligan} voxelizes molecules into atomic density grids and employs a conditional VAE to generate 3D molecules. Luo et al. \citep{2021sbdd}, Peng et al. \citep{2022pocket2mol}, Liu et al. \citep{2022graphbp}, Zhang et al. \citep{ zhang2023molecule, zhang2023learning} adopt autoregressive models that sequentially place atoms or molecular fragments within target binding site. Recently, non-autoregressive generative frameworks, including diffusion models \citep{targetdiff, guan2024decompdiff, qian2024kgdiff, gu2024aligning, zhou2025multi}, flow matching \citep{zhang2024flexsbdd, zhou2025prior}, and Bayesian flow networks \citep{qu2024molcraft, qiu2025empower}, have also been extensively applied to SBDD. These methods iteratively refine all atoms at each step and have demonstrated promising performance. By comparison, most existing work on dual-target molecular generation adopts sequence-based approaches. Isigkeit et al. \citep{isigkeit2024automated},  Srinivasan et al. \citep{srinivasan2024generation} fine-tune chemical language models toward the chemical space of dual-target compounds to generate SMILES sequences. Munson et al. \citep{munson2024novo}, Chen et al. \citep{chen2024structure} introduce reinforcement learning systems that score generated sequences based on their predicted interaction capabilities with two targets. In a recent study, Zhou et al. \citep{zhou2024reprogramming} propose reprogramming a pretrained target-aware diffusion model for the dual-target setting in a zero-shot manner. However, the proposed framework still suffers from insufficient integration of dual-target information and the suboptimal alignment algorithm between the two targets. In this work, we introduce the Fused Bayesian flow network together with improved alignment strategies to address these issues.

\paragraph{Bayesian Flow Networks}

The Bayesian Flow Networks, proposed by Graves et al. \citep{graves2023bayesian}, represent a new class of generative models that combines Bayesian inference with flow-based modeling. 
Different from diffusion models and most other probabilistic networks, which learn a mapping from data to a distribution, BFNs instead embody a function that transforms one distribution into another.
A notable advantage of this formulation is that the generative process is fully continuous and differentiable—properties not inherent to discrete data types (e.g., atom types)—thereby expanding the applicability of BFNs to discrete domains. Moreover, Song et al. \citep{song2024unified} shows that BFNs exhibit a more favorable inductive bias than diffusion models, making it better suited for noise-sensitive data such as molecular geometries. Since being proposed, this framework has attracted considerable attention and has demonstrated promising potential in various fields such as computer vision \citep{dou2025image, zheng2025target, penglinknowledge, zhangcontrollable} and biomolecule design \citep{qu2024molcraft,  song2024unified, atkinson2025protein, qiu2024empower, qian2025full}. Inspired by BFNs' mechanism of operating on the parameters of distributions, we propose a fused Bayesian flow network to perform distribution fusion in the parameter space, allowing the dual-target constraint to continuously participate in the Bayesian update throughout the generation process.

\section{Preliminaries}

\paragraph{Notations}


A target protein $\mathcal{P}$ is represented as a set of atoms $\mathcal{P}={(\mathbf{x}_P^{(i)},\mathbf{v}_P^{(i)})}_{i=1}^{N_P}$, where $\mathbf{x}_P^{(i)} \in \mathbb{R}^3$ denotes the 3D coordinates of the $i$-th protein atom, and $\mathbf{v}_P^{(i)} \in \mathbb{R}^{N_f}$ is a one-hot vector encoding its features (e.g., element type and amino acid type). Here, $N_P$ represents the number of atoms in the protein and $N_f$ is the feature dimension of each protein atom. Similarly, a binding molecule is defined as $\mathcal{M}={(\mathbf{x}_M^{(i)},\mathbf{v}_M^{(i)})}_{i=1}^{N_M}$, where $N_M$ is the number of atoms, $\mathbf{x}_M^{(i)} \in \mathbb{R}^3$ denotes the 3D coordinates of the $i$-th atom, and $\mathbf{v}_M^{(i)} \in \mathbb{R}^{K}$ represents its atom type, with $K$ being the molecule atom type dimension. The molecular representation can be simplified as $\mathbf{m}=[\mathbf{x}_M, \mathbf{v}_M]$, where $\mathbf{x}_M \in \mathbb{R}^{N_M \times 3}$ and $\mathbf{v}_M \in \mathbb{R}^{N_M \times K}$, and $[\cdot,\cdot]$ denotes the concatenation operator. Correspondingly, the binding pocket is represented as $\mathbf{p}=[\mathbf{x}_P, \mathbf{v}_P]$, where $\mathbf{x}_P \in \mathbb{R}^{N_P \times 3}$ and $\mathbf{v}_P \in \mathbb{R}^{N_P \times N_f}$.

\paragraph{Problem Formulation}
Dual-target drug design aims to generate ligand molecules that can simultaneously bind to two target proteins. The task can be formulated as a conditional generative model by $p(\mathbf{m}|\mathcal{T}\mathbf{p}_1, \mathbf{p}_2)$, where a transformation operator $\mathcal{T}$ is introduced. Since protein binding pockets exhibit diverse geometric shapes and chemical characteristics, spatial alignment between the two pockets is required when modeling the conditional distribution given both pockets \citep{zhou2024reprogramming}. The transformation $\mathcal{T}$ includes a translation $\mathcal{T}_T$ and a rotation $\mathcal{T}_R$, and the transformed pocket is given by $\mathcal{T P}_1 = (\mathcal{T} _T \circ \mathcal{T} _R)\mathcal{P}_1 =\{(\mathbf{R} \mathbf{x} ^{(i)}_{P_1}+\mathbf{t} ), \mathbf{v} _{P_1}^{(i)}\}^{N_{P_1}} _{i=1}$, where $\mathbf{R}$ is the rotation matrix and $\mathbf{t}$ is the translation vector.

\paragraph{Molecular Design via BFNs}

BFNs formulate the generative process as messages exchange between a sender and a receiver. At timestep $t_i$, the sender perturbs the molecule according to a predefined noise schedule $\beta(t_i)$ to build the \textit{sender distribution} $p_S(\mathbf{y}_i|\mathbf{m}, \mathbf{p};\alpha_i)$, from which a latent variable $\mathbf{y}_i$ is sampled and sent to the receiver. Here, $\alpha_i$
 denotes the noise factor in the schedule $\beta(t_i)$. The receiver then feeds the parameters $\boldsymbol{\theta}$ of the \textit{input distribution} $p_I(\mathbf{m}| \mathbf{p}, \boldsymbol{\theta})$ into a neural network $\boldsymbol{\Psi}$, which outputs an estimate of the original molecule, yielding in an \textit{output distribution} $p_O$. Noise at the same level as the sender is then added to $p_O$ to obtain the predicted latent, thereby constructing the
\textit{receiver distribution} $p_R$:
\begin{align}
p_R(\mathbf{y}_i|\boldsymbol{\theta}_{i-1}, \mathbf{p};t_{i}) = \mathop{\mathbb{E}}\limits_{\hat{\mathbf{m}}\sim p_O}p_S(\mathbf{y}_i|\hat{\mathbf{m}};\alpha_i), \quad&
    p_O(\hat{\mathbf{m}}|\boldsymbol{\theta}_{i-1}, \mathbf{p};t_{i-1}) = \boldsymbol{\Psi}(\boldsymbol{\theta}_{i-1}, \mathbf{p},t_{i-1}).
\end{align}

Unlike diffusion models that operate directly on noisy latent, BFN updates the parameters $\boldsymbol\theta$ via the Bayesian update function $h$ derived from Bayesian inference rules. Provided that the input distribution independently models all variables in the data, the resulting \textit{Bayesian update distribution} $p_U$ admits a closed-form expression while maintaining fully continuous parameters:
\begin{equation}
p_U(\boldsymbol\theta_i|\boldsymbol\theta_{i-1}, \mathbf{m}, \mathbf{p};\alpha_i)=\mathop{\mathbb{E}}\limits_{\mathbf{y}_i\sim p_S}\delta \big(  \boldsymbol{\theta}_i-h(\boldsymbol{\theta}_{i-1},\mathbf{y}_i,\alpha_i) \big),\label{p_U}
\end{equation}
where $\delta(\cdot)$ is Dirac delta distribution. Due to the additivity of accuracy \citep{graves2023bayesian}, the \textit{Bayesian flow distribution} $p_F$ can be derived as the marginal distribution of $p_U$, accounting for all possible intermediate updates from $t_0$ to $t_i$:
\begin{align}
p_F(\boldsymbol\theta_i| \mathbf{m}, \mathbf{p};t_i) = \mathop{\mathbb{E}}\limits_{\boldsymbol\theta_{1\dotsi-1}\sim p_U}p_U(\boldsymbol\theta_i|\boldsymbol\theta_{i-1}, \mathbf{m}, \mathbf{p};\alpha_i)=p_U(\boldsymbol\theta_i|\boldsymbol\theta_0, \mathbf{m}, \mathbf{p};\beta(t_i)).
\end{align}
After each update, the receiver again feeds the parameters of the input distribution into $\boldsymbol{\Psi}$ which outputs the parameters of $p_O$. This process is repeated for $n$ steps until the receiver can predict the molecule with sufficient accuracy, allowing the sender to transmit it without noise.



\begin{figure}[htbp]
\vspace{-0.7em}
  \centering
  \includegraphics[width=1\linewidth]{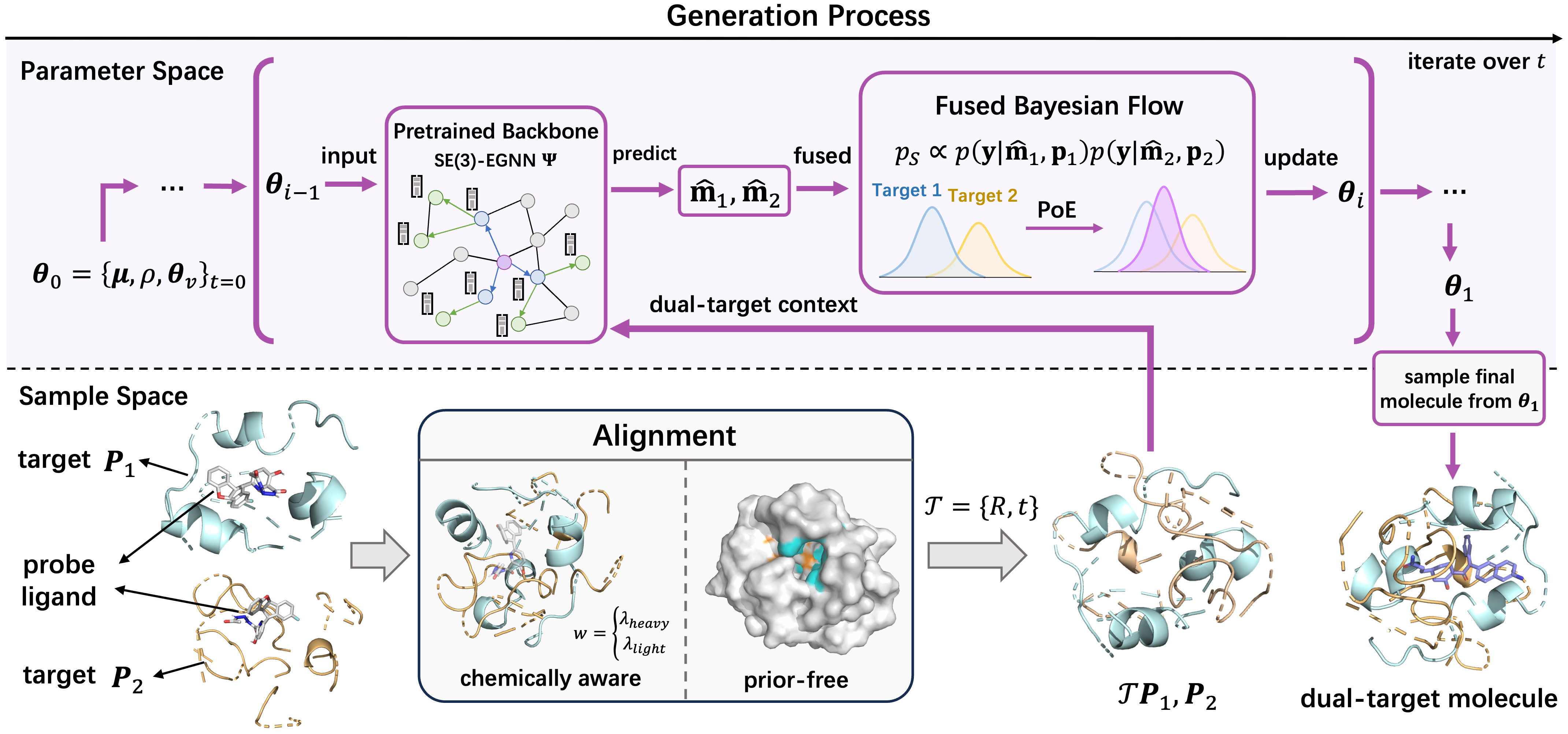}
  \caption{Overview of FusedBFN for dual-target molecular generation. We first aligns two protein pockets using either a chemically aware prior-based strategy or a prior-free alignment method. A shared pretrained SE(3)-equivariant network then predicts target-specific molecular estimates from the current parameter state. By fusing information from dual-target, FusedBFN progressively concentrates the parameter distribution toward molecular regions jointly supported by both binding pockets. The final molecule is sampled from the terminal parameter state and is expected to interact with both targets simultaneously. }
  \label{overview}
  \vspace{-1.5 em}
\end{figure}

\section{Method} 

This section elaborates on the implementation details of FusedBFN. First, we formulate dual-target molecular generation as a fused Bayesian flow in the continuous parameter space, where the distributions conditioned on two protein pockets are integrated. Based on this framework, we derive the fused Bayesian flow for continuous atom coordinates and discrete atom types, while preserving SE(3)-equivariance throughout parameter-space sampling. Additionally, we introduce a chemically aware alignment strategy together with a prior-free pocket alignment method.

\subsection{Fused Bayesian Flow
Networks} 
BFN views the generative process as message exchanging between a sender and a receiver, where the sender first perturbs the molecule with noise before transmission. We therefore perform the fusion of the two Bayesian flows starting from the sender distribution. Inspired by multimodal conditional image synthesis \citep{kutuzova2021multimodal,huang2022multimodal}, 
the sender distribution can be modeled using a product-of-experts (PoE) \citep{hinton2002training} formulation, which naturally amplifies molecular patterns supported by dual targets while suppressing modes that are only compatible with one target:
\begin{equation}
p_S(\mathbf{y}|\mathbf{m}_1, \mathbf{m}_2, \mathcal{T}\mathbf{p}_1, \mathbf{p}_2;\alpha) \propto p(\mathbf{y}|\mathbf{m}_1,\mathcal{T}\mathbf{p}_1;\alpha)p(\mathbf{y}| \mathbf{m}_2,\mathbf{p}_2;\alpha).
\label{fused sender}
\end{equation}

\paragraph{Continuous Data}
Following previous work \citep{hoogeboom2022equivariant,qu2024molcraft}, the continuous atom coordinates are modeled as a Gaussian distribution $\mathcal{N}(\mathbf{x}|\boldsymbol{\mu}, \rho^{-1}\mathbf{I})$, with parameters $\boldsymbol\theta^x=\{\boldsymbol{\mu}, \rho\}$, where $\boldsymbol\mu$ is learned and $\rho$ is specified by the noise factor $\alpha$. The prior $\boldsymbol\theta_0^x$ is set as a standard Gaussian \citep{graves2023bayesian} and the corresponding Bayesian update function $h(\{\boldsymbol{\mu}_{i-1}, \rho_{i-1}\}, \mathbf{y}^x, \alpha^x)=\{\boldsymbol{\mu}_i, \rho_i\}$ is defined as:
\begin{align}
    \rho_i = \rho_{i-1}+\alpha^x_i, \quad& \boldsymbol{\mu}_i = \frac{\boldsymbol{\mu}_{i-1}\rho_{i-1}+\mathbf{y}^x\alpha_i^x}{\rho_i}.
    \label{h of x}
\end{align}
In the single-target case, the sender distributions of atom coordinates conditioned on $\mathcal{P}_1$
 and $\mathcal{P}_2$ are respectively defined as follows:
\begin{align}
p_S(\mathbf{y}^x|\mathbf{x}_1, \mathbf{p}_1)=\mathcal{N}(\mathbf{y}^x|\mathbf{x}_1, (\alpha^x)^{-1}\mathbf{I}), \quad& p_S(\mathbf{y}^x|\mathbf{x}_2, \mathbf{p}_2)=\mathcal{N}(\mathbf{y}^x|\mathbf{x}_2, (\alpha^x)^{-1}\mathbf{I}).
\end{align}
Then, according to Eq. \ref{fused sender}, the fused distribution is given by:
\begin{equation}
p_S(\mathbf{y}^x|\mathbf{x}_1, \mathbf{x}_2, \mathcal{T}\mathbf{p}_1, \mathbf{p}_2) \propto \mathcal{N}(\mathbf{y}^x|\frac{\mathbf{x}_1+\mathbf{x_2}}{2}, \frac{1}{2}(\alpha^x)^{-1}\mathbf{I}).\label{fused sender of x}
\end{equation}
By substituting Eq. \ref{h of x} and Eq. \ref{fused sender of x} into Eq. \ref{p_U}, we obtain the Bayesian update distribution of the fused parameters $\boldsymbol{\theta}^x$:
\begin{equation}
p_U(\boldsymbol{\theta}_i^x|\boldsymbol{\theta}_{i-1}^x, \mathbf{x}_1, \mathbf{x}_2, \mathcal{T}\mathbf{p}_1, \mathbf{p}_2;\alpha^x_i) = \mathcal{N}(\boldsymbol{\mu}_i|\frac{\alpha^x_i(\mathbf{x}_1+\mathbf{x} _2)}{2\rho_i}+\frac{\boldsymbol{\mu}_{i-1}\rho_{i-1}}{\rho_i},\frac{\alpha^x_i}{2\rho_i^2}\mathbf{I} ).
\label{p_U of x}
\end{equation}
To derive the fused Bayesian flow distribution $p_F$ of $\boldsymbol{\theta}^x$, we present the following proposition (the full proof is given in Appendix \ref{Proof of Proposition 1}).

\textbf{Proposition 1.} \textit{In the fused Bayesian flow for continuous variables, the sender accuracies are additive, i.e.,}
$\mathop{\mathbb{E}}\limits_{p_U(\boldsymbol{\theta}^x_{i-1}|\boldsymbol{\theta}^x_{i-2}, \cdot;\alpha_{i-1})}p_U(\boldsymbol{\theta}^x_i|\boldsymbol{\theta}^x_{i-1},\cdot;\alpha_i)=p_U(\boldsymbol{\theta}_i^x|\boldsymbol{\theta}_{i-2}^x, \cdot;\alpha_i+\alpha_{i-1}).$

Proposition 1 indicates that for continuous coordinates, the proposed fusion does not break the closed-form Bayesian update of BFNs. Thus, $p_F$ takes the following form:
\begin{align}
p_F(\boldsymbol{\theta}^x|\mathbf{x}_1, \mathbf{x}_2, \mathcal{T}\mathbf{p}_1, \mathbf{p}_2;t_i) &= \mathop{\mathbb{E}}\limits_{\boldsymbol{\theta}_{1\cdots i-1}^x\sim p_U}[p_U(\boldsymbol{\theta}^x_i|\boldsymbol{\theta}^x_{i-1}, \mathbf{x}_1, \mathbf{x}_2, \mathcal{T}\mathbf{p}_1, \mathbf{p}_2;\alpha^x_i)] \notag \\ 
&= p_U(\boldsymbol{\theta}^x_i|\boldsymbol{\theta}^x_0, \mathbf{x}_1, \mathbf{x}_2, \mathcal{T}\mathbf{p}_1, \mathbf{p}_2;\beta^x(t_i)) \notag \\ 
&= \mathcal{N}(\boldsymbol{\mu}_i|\frac{\beta^x(t_i)}{1+\beta^x(t_i)}\frac{\mathbf{x}_1+\mathbf{x}_2}{2}, \frac{\beta^x(t)}{2(1+\beta^x(t_i))^2}\mathbf{I}), \label{p_F of x}
\end{align}
where $\beta^x (t_i) = \int_{t'=0}^{t_i} \alpha^x(t') dt'$, $\rho_i = 1+\beta^x(t_i)$. Denote $\gamma^x(t_i):\overset{\text{def}}{=}\frac{\beta^x(t_i)}{1+\beta^x(t_i)}$, then Eq. \ref{p_F of x} can be simplified as:
\begin{align}
p_F(\boldsymbol{\theta}^x|\mathbf{x}_1, \mathbf{x}_2, \mathcal{T}\mathbf{p}_1, \mathbf{p}_2;t_i) &= \mathcal{N}(\boldsymbol{\mu}_i|\gamma^x(t_i)\frac{\mathbf{x}_1+ \mathbf{x}_2}{2}, \frac{1}{2}\gamma^x(t_i)(1-\gamma^x(t_i))\mathbf{I}).
\end{align}

\paragraph{Discrete data}
For discrete atom types, we model them using a categorical distribution parameterized by learnable parameters $\boldsymbol{\theta}^v \in \mathbb{R}^{N_M\times K}$, with the prior $\boldsymbol{\theta}_0$ set to a uniform distribution following \citep{graves2023bayesian}. The Bayesian update function is given by:
\begin{equation}
 h(\boldsymbol{\theta}^v_{i-1}, \mathbf{y}^v, \alpha^v_i):\overset{\text{def}}{=} \frac{e ^{\mathbf{y}^v} \boldsymbol{\theta}_{i-1}}{ {\textstyle \sum_{k=1}^{K}}e ^{\mathbf{y}^v_k}(\boldsymbol{\theta}_{i-1})_k }.
 \label{h of v}
\end{equation}
The sender distributions over atom types in the single-target setting under conditions $\mathcal{P}_1$ and $\mathcal{P}_2$ are defined as follows, respectively:
\begin{align}
p_S(\mathbf{y}^v|\mathbf{v}_1, \mathbf{p}_1; \alpha^v) = \mathcal{N} (\mathbf{y}^v|\alpha^v(K\mathbf{e}_{\mathbf{v}_1}-1),\alpha^vK\mathbf{I}), \quad& p_S(\mathbf{y}^v|\mathbf{v}_2, \mathbf{p}_2; \alpha^v) = \mathcal{N} (\mathbf{y}^v|\alpha^v(K\mathbf{e}_{\mathbf{v}_2}-1),\alpha^v K\mathbf{I}),
\label{p_s of v}
\end{align}
where $\mathbf{e}_{\mathbf{v}_r}=[\mathbf{e}_{\mathbf{v}_r^{(1)}}, ...,\mathbf{e}_{\mathbf{v}_r^{(N_M)}}]\in \mathbb{R}^{N_M \times K}$, $r\in{1,2}$, $\mathbf{e}_j \in \mathbb{R}^K$ is the projection from the class index $j$ to the length-$K$ one-hot vector. Substituting Eq. \ref{p_s of v} into Eq. \ref{fused sender} yields the PoE form:
\begin{equation}
p_S(\mathbf{y}^v|\mathbf{v}_1,\mathbf{v}_2, \mathcal{T}\mathbf{p}_1, \mathbf{p}_2; \alpha^v)\propto \mathcal{N}(\mathbf{y}^v|\frac{\alpha^v}{2}(K\mathbf{e_{\mathbf{v}_1}}+K\mathbf{e_{\mathbf{v}_2}}-2), \frac{\alpha^v}{2} K\mathbf{I}).
\label{fused p_s of v}
\end{equation}
Plugging Eqs. \ref{h of v} and \ref{fused p_s of v} in Eq. \ref{p_U} gives the fused Bayesian update for $\boldsymbol{\theta}^v$:
\begin{equation}
p_U(\boldsymbol{\theta}^v_i|\boldsymbol{\theta}^v_{i-1}, \mathbf{v}_1, \mathbf{v}_2, \mathcal{T}\mathbf{p}_1, \mathbf{p}_2;\alpha^v) = \mathop{\mathbb{E}}\limits_{\mathcal{N}(\mathbf{y}^v|\frac{\alpha^v}{2}(K\mathbf{e}_{\mathbf{v}_1}+K\mathbf{e}_{\mathbf{v}_2}-2), \frac{\alpha^v}{2}K\mathbf{I})}[\delta(\boldsymbol{\theta}^v-\frac{\mathbf{e}^{\mathbf{y}^v} \boldsymbol{\theta}^v_{i-1}}{ {\textstyle \sum_{k=1}^{K}}\mathbf{e} ^{\mathbf{y}^v_k}(\boldsymbol{\theta}^v_{i-1})_k })].
\end{equation}
The additivity of sender accuracies still holds for discrete variables under the fused Bayesian flow (see detailed proof in Appendix \ref{Proof of Proposition 2}).

\textbf{Proposition 2.} \textit{For discrete variables, the fused Bayesian flow satisfies the same additive property of accuracies, i.e.,}
$\mathop{\mathbb{E}}\limits_{p_U(\boldsymbol{\theta}^v_{i-1}|\boldsymbol{\theta}^v_{i-2}, \cdot;\alpha^v_{i-1})}p_U(\boldsymbol{\theta}^v_i|\boldsymbol{\theta}^v_{i-1},\cdot;\alpha^v_i)=p_U(\boldsymbol{\theta}^v_i|\boldsymbol{\theta}^v_{i-2}, \cdot;\alpha^v_i+\alpha^v_{i-1}).$

Proposition 2 ensures that atom types from two pockets can be fused consistently in the categorical parameter space, which is essential for jointly modeling dual-target distribution. Then, $p_F$ as the marginal distribution over the parameters at time $t_i$ is given as:
\begin{align}
p_F(\boldsymbol{\theta}^v|\mathbf{v}_1,\mathbf{v}_2, \mathcal{T}\mathbf{p}_1, \mathbf{p}_2; t_i) &= \mathop{\mathbb{E}}\limits_{\boldsymbol{\theta}_{1\cdots i-1}^v\sim p_U}p_U(\boldsymbol{\theta}^v_i|\boldsymbol{\theta}^v_{i-1}, \mathbf{v}_1, \mathbf{v}_2, \mathcal{T}\mathbf{p}_1, \mathbf{p}_2;\alpha^v_i)  \\
&= p_U(\boldsymbol{\theta}^v_i|\boldsymbol{\theta}^v_0, \mathbf{v}_1, \mathbf{v}_2, \mathcal{T}\mathbf{p}_1, \mathbf{p}_2;\beta^v(t_i)) \\
&= \mathop{\mathbb{E}}\limits_{\mathcal{N}(\mathbf{y}^v|\frac{\beta^v(t_i)}{2}(K\mathbf{e}_{\mathbf{v}_1}+K\mathbf{e}_{\mathbf{v}_2}-2), \frac{\beta^v(t_i)}{2}K\mathbf{I})}[\delta(\boldsymbol{\theta}^v-\frac{\mathbf{e}^{\mathbf{y}^v} \boldsymbol{\theta}^v_{i-1}}{ {\textstyle \sum_{k=1}^{K}}\mathbf{e} ^{\mathbf{y}^v_k}(\boldsymbol{\theta}^v_{i-1})_k })].
\end{align}
Since the prior is uniform with $\boldsymbol{\theta}_0=\mathbf{\frac{1}{K}}$, this can be simplified as:
\begin{equation}
p_F(\boldsymbol{\theta}^v|\mathbf{v}_1,\mathbf{v}_2, \mathcal{T}\mathbf{p}_1, \mathbf{p}_2; t_i) = \mathop{\mathbb{E}}\limits_{\mathcal{N}(\mathbf{y}^v|\frac{\beta^v(t_i)}{2}(K\mathbf{e}_{\mathbf{v}_1}+K\mathbf{e}_{\mathbf{v}_2}-2), \frac{\beta^v(t_i)}{2}K\mathbf{I})}[\delta(\boldsymbol{\theta}^v-\text{softmax}(\mathbf{y}^v))].
\end{equation}

\paragraph{Sampling Process}
To avoid introducing excessive noise, we adopt the parameter-space sampling strategy proposed in \citep{qu2024molcraft}, in which information flows as $ \boldsymbol\theta_{i-1} \overset{\boldsymbol\Psi}{\rightarrow} (\hat{\mathbf{m} } _1, \hat{\mathbf{m} }_2) \overset{p_F}{\rightarrow} \boldsymbol\theta_{i} $, thereby bypassing the noisy data sampling required by Bayesian updating. The estimated $(\hat{\mathbf{m}}_1, \hat{\mathbf{m}}_2)$ are used in $p_F$ to directly update the parameters at the next step, where $\hat{\mathbf{m}}_1 = [\mathbf{\hat{x}}_1, \mathbf{\hat{v}}_1] \sim p_O(\boldsymbol{\theta},\mathcal{T}\mathbf{p}_1,t)$,  $\hat{\mathbf{m}}_2 = [\mathbf{\hat{x}}_2, \mathbf{\hat{v}}_2] \sim p_O(\boldsymbol{\theta},\mathbf{p}_2,t)$ and $p_O$ is parameterized by $\boldsymbol{\Psi}$.
Here, 
$\boldsymbol{\Psi}$ is an SE(3)-equivariant GNN, which ensures that the generative process remains invariant to translations and rotations of the protein-ligand complex, an important inductive bias for 3D molecular generation \citep{hoogeboom2022equivariant, kohler2020equivariant,garcia2021n,xu2022geodiff}. 

\textbf{Proposition 3.}
\textit{The generative process preserves SE(3)-equivariant if $\boldsymbol{\Psi}$ is parameterized by an SE(3)-equivariant network and the Center of Mass (CoM) of the pair $(\mathcal{T}\mathbf{p}_1, \mathbf{p}_2)$ is shifted to zero.}

This guarantees that the proposed fusion does not destroy the geometric inductive bias. The proof of Proposition 3 can be found in Appendix \ref{Proof of Proposition 3} and the sampling procedures is summarized in the Appendix \ref{algrithm}. $\boldsymbol{\Psi}$ uses shared node parameters for both $\mathbf{p}_1$ and $\mathbf{p}_2$. Although $\boldsymbol{\Psi}$ is pretrained on the single-target setting due to the scarcity of data resources, the results in Section \ref{Main Results} indicate that the information fusion in the fused Bayesian flow effectively extends it to the dual-target scenario.

\subsection{Alignment}
\label{Alignment}


Protein pockets often exhibit highly irregular local geometries, and substantial differences in residue distribution, chemical properties, and surface contours across different targets. Therefore, directly performing alignment based only on pocket structures makes it difficult to obtain a stable correspondence that is meaningful for molecular binding. To alleviate this issue, previous work \citep{zhou2024reprogramming} leverages protein–ligand binding priors by first docking a probe ligand to $\mathcal{P}_1$ and $\mathcal{P}_2$, respectively, and then aligning the two binding poses of the ligand to obtain the transformation matrix. Following this idea, we further improve the prior-based alignment strategy by considering the chemical semantics of different atoms. Specifically, different weights are assigned to atoms according to their types when aligning the binding priors, with larger weights given to heavy atoms: 
\begin{equation}
\label{eq:heavy_atom_weight}
w_i =
\begin{cases}
\lambda_{\mathrm{heavy}}, & \text{if atom } i \text{ is a heavy atom},\\[3pt]
\lambda_{\mathrm{light}}, & \text{otherwise},
\end{cases}
\qquad
\lambda_{\mathrm{heavy}} > \lambda_{\mathrm{light}} > 0.
\end{equation}
The motivation is that heavy atoms usually form the main topological scaffold of the ligand and provide more distinctive geometric constraints, whereas light atoms contribute less to transformation estimation and are more likely to introduce local perturbations. In FusedBFN, this chemically aware strategy is used as the default alignment method.

To provide a simpler option, we further propose a direct pocket alignment method that does not require priors. Since ligand binding is primarily governed by the geometry and physicochemical properties of the pocket surface \citep{zhang2011identification}, we recast protein alignment as the alignment of the surface-atom point clouds of the two pockets. Specifically, we first use P2Rank \citep{krivak2018p2rank} to localize the binding surface and extract the corresponding surface atoms. We then perform coarse alignment between the two surface point clouds using RANSAC \citep{fischler1981random}, followed by ICP \citep{besl1992method} for further refinement, yielding the final alignment result. It is worth noting that, in the absence of explicit binding information, directly aligning two pockets based solely on their structures is inherently more challenging. Even so, our method still achieves performance close to that of the prior-dependent approach (see Section \ref{Ablation Studies}).

\section{Experiments}

\subsection{Experimental Setup}

\paragraph{Dataset}

We evaluate our method on the dual-target benchmark proposed in \citep{zhou2024reprogramming}. This dataset is curated from synergistic drug combinations rather than arbitrary target pairs, making the selected target pairs more relevant to practical drug discovery. It contains 12,917 target pairs involving 438 unique drugs, and each target is associated with a reference ligand. Following the protocol of \citep{zhou2024reprogramming}, all target pairs are used for evaluation.

\paragraph{Baselines}

FusedBFN is compared with various baseline methods: \textbf{TargetDiff} \citep{targetdiff} is a diffusion-based model that generate 3D molecules in a non-autoregressive manner. \textbf{MolCRAFT} \citep{qu2024molcraft} adopts the Bayesian Flow Network framework and performs sampling in a continuous parameter space. Both methods are designed for the single-target SBDD task.
\textbf{CompDiff} and \textbf{DualDiff} are two dual-target diffusion frameworks proposed in \citep{zhou2024reprogramming}. CompDiff composes the reverse drifts predicted from two target contexts, while DualDiff further composes SE(3)-equivariant messages at each layer of the network and has been shown to outperform CompDiff on the dual-target benchmark.

\begin{table}[htbp]
\vspace{-1em}
\begin{center}
\captionsetup{skip=4pt}
\caption{Summary of the binding affinity and molecular properties of reference ligands and molecules generated by FusedBFN and other baselines in the dual-target setting. $(\uparrow) / (\downarrow)$ indicates that higher/lower values are preferred. Top 2 results are highlighted with \textbf{bold text} and \ul{underlined text}, respectively.
}
\renewcommand{\arraystretch}{1.3}
\label{tab1}
\resizebox{1\columnwidth}{!}{
\begin{tabular}{c|cc|cc|cc|cc|cc|cc|cc}
\toprule
\multirow{2}{*}{Methods} & \multicolumn{2}{c|}{P-1 Vina Dock $(\downarrow)$} & \multicolumn{2}{c|}{P-2 Vina Dock $(\downarrow)$} & \multicolumn{2}{c|}{Max Vina Dock $(\downarrow)$} & \multicolumn{2}{c|}{Dual High Aff. $(\uparrow)$} & \multicolumn{2}{c|}{QED $(\uparrow)$} & \multicolumn{2}{c|}{SA $(\uparrow)$} & \multicolumn{2}{c}{Diversity $(\uparrow)$} \\
                         & Avg.                    & Med.                    & Avg.                    & Med.                    & Avg.                    & Med.                    & Avg.                    & Med.                   & Avg.              & Med.              & Avg.              & Med.             & Avg.                 & Med.                \\ \midrule
Reference                      & -7.67                   & -7.83                   & -4.90                   & -7.33                   & -4.37                   & -7.10                   & -                       & -                      & 0.53              & 0.55              & 0.74              & 0.77             & -                    & -                   \\ \midrule
TargetDiff               & -8.62                   & -8.61                   & -6.89                   & -7.68                   & -6.56                   & -7.39                   & 44.6\%                  & 42.9\%                 & 0.50              & 0.51              & 0.58              & 0.58             & \ul{0.70}           & \ul{0.71}          \\
MolCRAFT                 & \textbf{-9.01}          & \textbf{-8.88}          & -7.60                   & -7.92                   & -7.30                   & -7.63                   & \ul{52.2\%}            & \ul{50.0\%}           & 0.52              & 0.53              & \ul{0.66}              & \ul{0.65}             & \ul{0.70}           & \ul{0.71}          \\
CompDiff                 &  -8.35                   & -8.48                  & -8.42             &   \ul{-8.53}            & -7.50             & -7.80             & 51.2\%                  & \ul{50.0\%}           & \ul{0.55}              & 0.56              & 0.59              & 0.59             & \textbf{0.72}        & \textbf{0.72}       \\
DualDiff                 & -8.38                    & -8.48                   &  \ul{ -8.43}                  &  -8.52                    & \ul{-7.60}               & \ul{-7.84}                   &  51.2\%                  & \ul{50.0\%}                  & \ul{0.55}              & \ul{0.57}              & 0.59              & 0.58             & 0.66        & 0.67       \\
FusedBFN                 & \ul{-8.74}              & \ul{-8.70}             & \textbf{-8.73}          & \textbf{-8.69}          & \textbf{-8.02}          & \textbf{-8.05}          & \textbf{57.8\%}         & \textbf{60.0\%}        & \textbf{0.56}        & \textbf{0.58}        & \textbf{0.69}        & \textbf{0.69}       & 0.69                 & 0.69                \\ \midrule
\end{tabular}
}
\end{center}
\vspace{-2em}
\end{table}

\begin{table}[htbp]
\vspace{-0.2em}
\begin{center}
\captionsetup{skip=4pt}
\caption{Evaluation results on molecular conformation stability under the dual-target setting. SE and Clash are computed using PoseCheck \citep{harris2023posecheck}. For SE, we report the 25th percentile, median, and 75th percentile.}
\renewcommand{\arraystretch}{1.2}
\label{tab2}
\resizebox{0.75\columnwidth}{!}{
\begin{tabular}{c|ccc|ccc|cc|cc}
\toprule
\multirow{2}{*}{Methods} & \multicolumn{3}{c|}{P-1 SE ($\downarrow$)} & \multicolumn{3}{c|}{P-2 SE ($\downarrow$)} & \multicolumn{2}{c|}{P-1 Clash ($\downarrow$)} & \multicolumn{2}{c}{P-2 Clash ($\downarrow$)} \\
                         & 25\%         & 50\%         & 75\%         & 25\%         & 50\%         & 75\%         & Avg.                  & Med.                  & Avg.                  & Med.                 \\ \midrule
Reference                      & 41           & 73           & 168          & 39           & 66           & 170          & 6.48                  & 4.00                   & 79.98                 & 46.00                \\ \midrule
TargetDiff               & 386          & 1266         & 11322        & 395          & 1214         & 11161        & \ul{9.93}            & \ul{7.00}            & 68.75                 & 47.00                \\
MolCRAFT                 & \ul{123}    & \ul{262}    & \ul{833}    & \ul{122}    & \ul{262}    & \ul{833}    & \textbf{6.08}         & \textbf{4.00}         & 66.68                 & 48.00                \\
CompDiff                 & 521          & 1671         & 17551        & 521          & 1671         & 17551        & 28.18                 & 21.00                 & 23.69           & 17.00          \\
DualDiff                 & 416          & 1112         & 9287        & 416          & 1112         & 9287        &  23.65                 & 16.00                 &  \ul{18.27}          & \ul{12.00}          \\
FusedBFN                 & \textbf{102} & \textbf{194} & \textbf{474} & \textbf{101} & \textbf{194} & \textbf{474} & 16.37                 & 10.00                 & \textbf{12.00}        & \textbf{7.00}        \\ \midrule
\end{tabular}
}
\end{center}
\vspace{-1.8em}
\end{table}

\paragraph{Evaluation metrics}

We use each method to generate 10 molecules for each pair of targets and evaluate generated molecules from the following aspects: 
(1) \textbf{Binding Affinity} to the two targets: For each target pair $(\mathcal{P}_1, \mathcal{P}_2)$, we assess the binding affinity using AutoDock Vina \citep{trott2010autodock}, following \citep{targetdiff, qu2024molcraft}.
Vina Dock performs a re-docking procedure to evaluate the optimal binding affinity.
We report \textbf{P-1 Vina Dock} and \textbf{P-2 Vina Dock}, which denote the Vina Dock scores of a generated molecule on $\mathcal{P}_1$ and $\mathcal{P}_2$, respectively. We further include \textbf{Max Vina Dock}, defined as the larger docking score between the two targets for each molecule. Since a dual-target molecule is expected to bind well to both targets simultaneously, a lower Max Vina Dock indicates better balanced dual-target binding. In addition, we report \textbf{Dual High Affinity}, which measures the proportion of generated molecules whose binding affinities exceed those of the reference molecules on both targets simultaneously. This directly reflects the success rate of achieving stronger binding to both targets at the same time. For reference ligands, TargetDiff and MolCRAFT, we evaluate them under the dual-target setting by generating molecules conditioned on $\mathcal{P}_1$ and then docking the resulting molecules to both $\mathcal{P}_1$ and $\mathcal{P}_2$. (2) \textbf{Molecular Properties}: we evaluate molecular properties using \textbf{QED} \citep{bickerton2012quantifying} for drug-likeness, \textbf{SA} \citep{ertl2009estimation} for synthesize accessibility, and \textbf{Diversity}. Following \citep{zhou2024reprogramming}, we summarize these metrics by reporting both the mean and the median over all generated molecules. (3) \textbf{Conformation Evaluation}: Strain energy (\textbf{P-1 SE} and \textbf{P-2 SE}) \citep{harris2023posecheck} measures the internal energy accumulated in a ligand due to conformational adjustments upon binding. Steric clashes (\textbf{P-1 Clash} and \textbf{P-2 Clash}) \citep{harris2023posecheck} quantify cases in which the distance between a protein atom and a ligand atom is smaller than the sum of their van der Waals radii, using a clash tolerance of $0.5$ $\mathring{\mathrm{A}}$. We further compute the \textbf{RMSD} \citep{zhou2024reprogramming} between the docked poses of each molecule on dual targets to assess binding mode consistency.

\begin{wrapfigure}{r}{7cm}
\vspace{-1em}
\centering
  \includegraphics[width=1\linewidth]{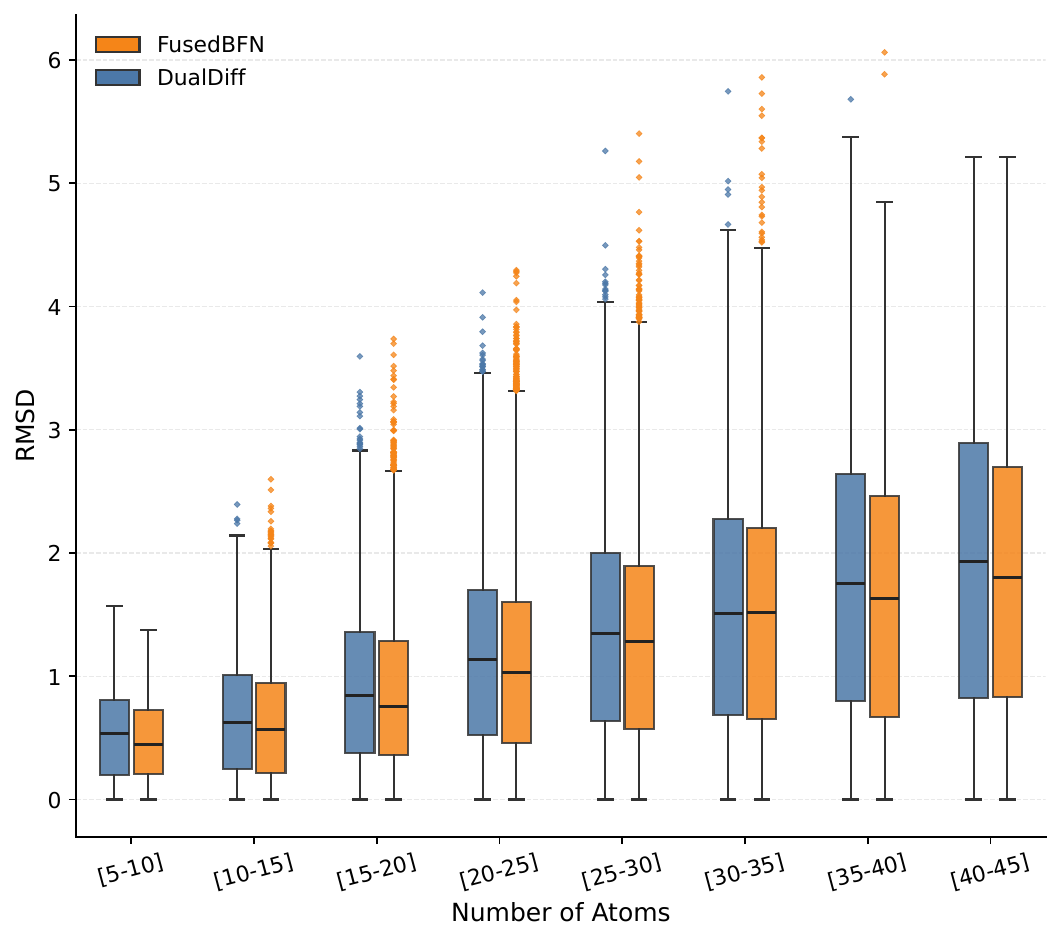}
  \caption{RMSD distributions of docked poses on dual targets across different atom-number intervals. Lower values indicate better cross-target pose consistency.}
  \label{rmsd}
  \vspace{-3em}
\end{wrapfigure}

\subsection{Main Results}
\label{Main Results}

\paragraph{Binding Affinity and Molecular Properties}

We evaluate all methods under the dual-target setting, with the results reported in Table \ref{tab1}. FusedBFN significantly outperforms other methods on binding-related metrics, and achieves the best performance on both Max Vina Dock and Dual High Affinity, indicating that the generated molecules exhibit high affinity to both targets simultaneously. TargetDiff and MolCRAFT are competitive methods for single-target drug design, yet their docking results drop markedly on $\mathcal{P}_2$. This observation suggests that single-target generation ability can not directly translate into effective dual-target design. Compared with the strongest baseline (i.e., DualDiff), FusedBFN not only yields better binding affinity ($p < 0.05$, see Section \ref{Significance Test on Affinity Metrics}) but also exhibits more favorable molecular properties, which highlights that information fusion in the continuous parameter space is more advantageous than drift in the mixed continuous-discrete sample space.

\paragraph{Conformation Evaluation}

\begin{figure}[htbp]
  \centering
\includegraphics[width=1\linewidth]{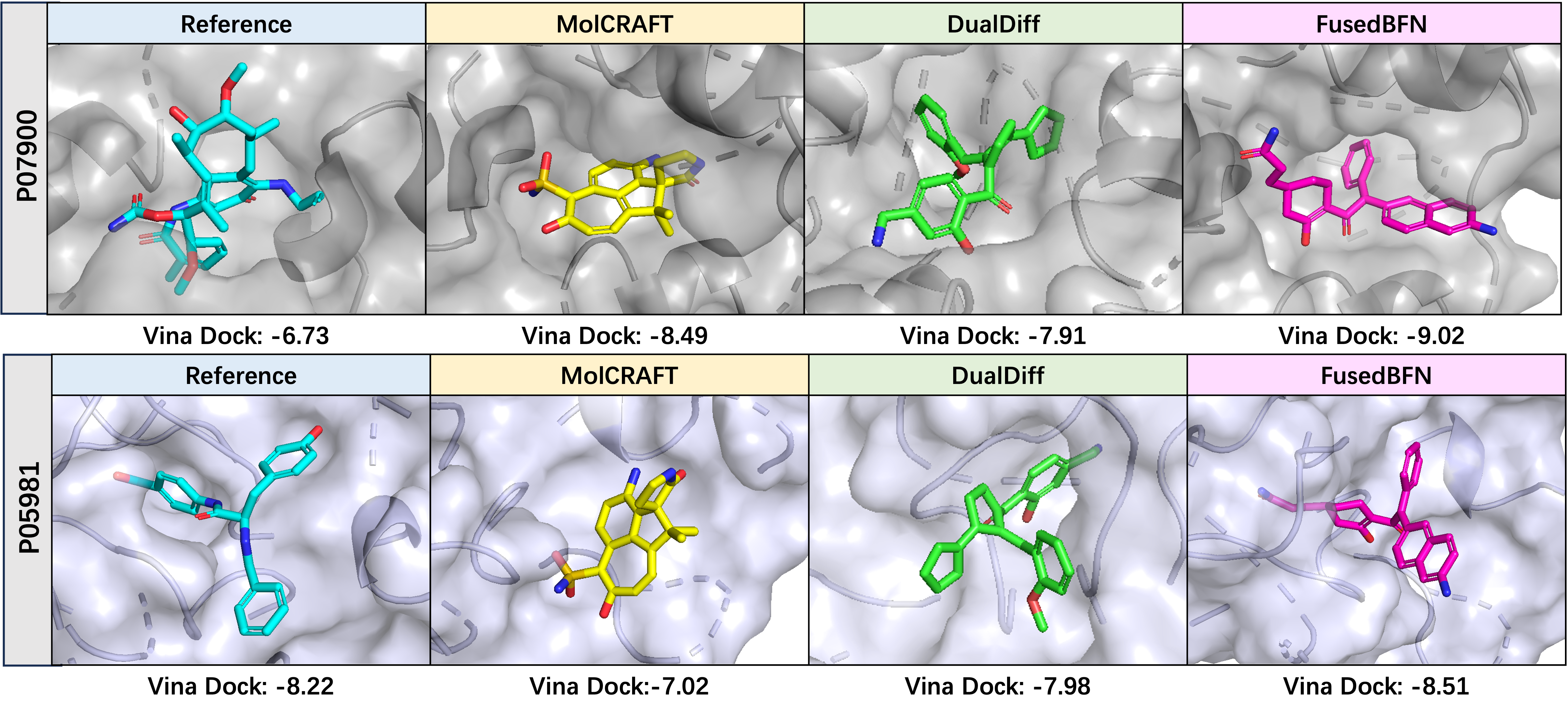}
  \caption{Visualizations of reference ligands and molecules generated by different methods for the dual-target pair (UniProt IDs: P07900, top; P05981, bottom).}
  \label{visualization}
  \vspace{-0.5em}
\end{figure}

We further evaluate the conformation stability of the generated molecules. As shown in Table \ref{tab2}, FusedBFN achieves the lowest strain energy on both targets, substantially surpassing all baselines. It also shows a clear advantage over DualDiff in terms of Clash, demonstrating the strong generative capability in the parameter space. As expected, TargetDiff and MolCRAFT yield lower Clash on $\mathcal{P}_1$ but perform poorly on $\mathcal{P}_2$, which further indicates that molecules generated for a single target do not readily generalize to another target. In addition, Figure \ref{rmsd} shows that FusedBFN yields lower RMSD than DualDiff across all atom-number ranges, suggesting that the molecules generated by FusedBFN can bind to both targets with smaller conformational changes. This means that, even without force field optimization or re-docking step, the molecules generated by FusedBFN are closer to the docking poses on dual targets and are therefore more reliable than those generated by DualDiff. Figure \ref{visualization} provides visualization of examples of molecules generated by different methods. 
For more examples, please refer to Appendix \ref{More Visualization Examples}.

\subsection{Ablation Studies}
\label{Ablation Studies}


\paragraph{Effect of Parameter Fusion} To validate the effectiveness of performing dual-target fusion in the parameter space, we compare the proposed fused Bayesian flow  (\textbf{P-Fused}) with a sample-space fusion variant (\textbf{S-Fused}), where the molecular estimates output by the shared network $\boldsymbol{\Psi}$ are averaged before parameter updating. We randomly extract 1,000 dual-target pairs from the dataset and generate 10 molecules for each pair. As shown in Table \ref{Ablation fusion}, P-Fused achieves stronger dual-target binding than S-Fused while maintaining more desirable molecular properties. These results indicate that fusion in the parameter space is more effective for dual-target molecular generation. Parameter-space fusion allows dual-target information to be continuously propagated during the Bayesian update process, whereas S-Fused cannot sufficiently model the dual-target constraints in the sample space.

\begin{table}[htbp]
\vspace{-1em}
\begin{center}
\captionsetup{skip=4pt}
\caption{Ablation study comparing sample-space and parameter-space fusion.}
\renewcommand{\arraystretch}{1.2}
\label{Ablation fusion}
\resizebox{1\columnwidth}{!}{
\begin{tabular}{c|cc|cc|cc|cc|cc|cc|cc}
\toprule
\multirow{2}{*}{Methods} & \multicolumn{2}{c|}{P-1 Vina Dock $(\downarrow)$} & \multicolumn{2}{c|}{P-2 Vina Dock $(\downarrow)$} & \multicolumn{2}{c|}{Max Vina Dock $(\downarrow)$} & \multicolumn{2}{c|}{Dual High Aff. $(\uparrow)$} & \multicolumn{2}{c|}{QED $(\uparrow)$} & \multicolumn{2}{c|}{SA $(\uparrow)$} & \multicolumn{2}{c}{Diversity $(\uparrow)$} \\
                         & Avg.                    & Med.                    & Avg.                    & Med.                    & Avg.                    & Med.                    & Avg.                    & Med.                   & Avg.              & Med.              & Avg.              & Med.             & Avg.                 & Med.                \\ \midrule
S-Fused                   & -8.42                   & -8.48                   & -8.51                   & -8.47                   & -7.78                   & -7.90                   & 49.8\%                  & 50.0\%                 & \textbf{0.54}     & 0.55              & 0.65              & 0.64             & \textbf{0.68}        & \textbf{0.68}       \\
P-Fused                     & \textbf{-8.81}          & \textbf{-8.78}          & \textbf{-8.84}          & \textbf{-8.78}          & \textbf{-8.15}          & \textbf{-8.16}          & \textbf{56.3\%}         & \textbf{60.0\%}        & \textbf{0.54}     & \textbf{0.56}     & \textbf{0.68}     & \textbf{0.68}    & \textbf{0.68}        & \textbf{0.68}       \\ \midrule
\end{tabular}
}
\end{center}
\vspace{-2em}
\end{table}

\paragraph{Different Strategies of Dual-Target Alignment} We design several variants based on FusedBFN to compare different alignment strategies. Here, \textbf{Pocket-RI} directly aligns the full protein pockets using RANSAC followed by ICP  refinement. \textbf{Surface-R} first extracts pocket-surface atoms with P2Rank and aligns them using RANSAC. \textbf{Surface-RI} further refines the RANSAC-based surface alignment with ICP. \textbf{LigPrior} first aligns the probe ligands and then transfers the resulting transformation to the corresponding protein pockets. \textbf{Lig-ChemPrior} further incorporates chemically aware weighting during ligand alignment. We randomly sample 1,000 dual-target pairs from the dataset and generate 10 molecules for each pair. The evaluation results are reported in Table \ref{Ablation alignment}. 

Lig-ChemPrior achieves the best overall performance, which shows that incorporating the chemical semantics of different atoms leads to a more stable alignment pattern. Surface-RI remains competitive even without interaction priors, offering a practical solution when reliable probe ligands are unavailable and providing a simplified alignment pipeline. It outperforms Pocket-RI because directly aligning full pockets is less effective when a large number of atoms are involved, which enlarges the alignment space and makes reliable registration more difficult. By focusing on the smaller set of pocket-surface atoms, Surface-RI yields more stable transformations. In addition, the improvement of Surface-RI over Surface-R further verifies the benefit of the ICP refinement stage.

\begin{table}[htbp]
\vspace{-1.2em}
\begin{center}
\captionsetup{skip=4pt}
\caption{Ablation study of different alignment strategies in FusedBFN.}
\renewcommand{\arraystretch}{1.2}
\label{Ablation alignment}
\resizebox{1\columnwidth}{!}{
\begin{tabular}{c|cc|cc|cc|cc|cc|cc|cc}
\toprule
\multirow{2}{*}{Methods} & \multicolumn{2}{c|}{P-1 Vina Dock $(\downarrow)$} & \multicolumn{2}{c|}{P-2 Vina Dock $(\downarrow)$} & \multicolumn{2}{c|}{Max Vina Dock $(\downarrow)$} & \multicolumn{2}{c|}{Dual High Aff. $(\uparrow)$} & \multicolumn{2}{c|}{QED $(\uparrow)$} & \multicolumn{2}{c|}{SA $(\uparrow)$} & \multicolumn{2}{c}{Diversity $(\uparrow)$} \\
                         & Avg.                    & Med.                    & Avg.                    & Med.                    & Avg.                    & Med.                    & Avg.                    & Med.                   & Avg.              & Med.              & Avg.              & Med.             & Avg.                 & Med.                \\ \midrule
Pocket-RI        & -8.47                   & -8.47                   & -8.46                   & -8.49                   & -7.73                   & -7.86                   & 49.4\%                  & 50.0\%           & 0.51              & 0.52              & 0.64              & 0.63             & 0.67                 & 0.67                \\
Surface-R              & -8.47                   & -8.49                   & -8.50                   & -8.50                   & -7.82                   & -7.90                   & 50.1\%                  & 50.0\%           & 0.54        & 0.55              & \textbf{0.68}     & \textbf{0.68}    & \textbf{0.70}        & \textbf{0.71}       \\
Surface-RI          & -8.50                   & -8.49                   & -8.56                   & -8.54                   & -7.86                   & -7.93                   & 50.7\%                  & 50.0\%           & 0.53              & 0.55              & 0.67        & 0.66       & 0.69           & 0.69          \\
LigPrior                 & -8.71             & -8.72             & -8.74             & -8.74             & -8.03             & -8.12             & 56.2\%            & 60.0\%       & \textbf{0.55}     & \textbf{0.57}     & \textbf{0.68}     & \textbf{0.68}    & 0.69           & 0.69          \\
Lig-ChemPrior            & \textbf{-8.81}          & \textbf{-8.78}          & \textbf{-8.84}          & \textbf{-8.78}          & \textbf{-8.15}          & \textbf{-8.16}          & \textbf{56.3\%}         & \textbf{60.0\%}        & 0.54        & 0.56        & \textbf{0.68}     & \textbf{0.68}    & 0.68                 & 0.68                \\ \midrule
\end{tabular}
}
\end{center}
\vspace{-2em}
\end{table}

\section{Conclusion}

In this work, we propose FusedBFN, a fused Bayesian Flow Network which formulates dual-target generation as distribution fusion in a unified continuous parameter space. During parameter updates, FusedBFN effectively propagates fused target-context information and enables the generation of molecules with simultaneous binding affinity toward both targets. Built upon a pretrained target-aware BFN backbone, the proposed framework extends knowledge learned from single-target data to the dual-target setting without additional training or fine-tuning. We further introduce a chemically aware dual-target alignment strategy, together with a prior-free pocket alignment method that provides a simpler alternative. Extensive experiments on the dual-target benchmark demonstrate that FusedBFN achieves superior dual-target binding performance while maintaining favorable molecular properties and conformation stability. Overall, FusedBFN provides a new perspective on dual-target drug design and holds promise for accelerating the discovery of dual-target therapeutics. In future work, we plan to extend the current framework to more general multi-target molecular generation scenarios.


{
\bibliographystyle{IEEEtran} \bibliography{neurips_2026}
}


\appendix
\newpage

\section{Algorithm}
\label{algrithm}

The sampling procedure of FusedBFN are summarized below.

\renewcommand{\algorithmicrequire}{\textbf{Input:}}
\renewcommand{\algorithmicensure}{\textbf{Output:}}

\begin{algorithm}[H]
    \caption{Sampling Procedure of FusedBFN}
    \label{train_cap}
    \setstretch{1.15}
    \begin{algorithmic}[1] 
    \Function{update}{$\hat{\mathbf{x}}_1, \hat{\mathbf{x}}_2 \in \mathbb{R}^{3N_M}$, $\hat{\mathbf{v}}_1, \hat{\mathbf{v}}_2 \in \mathbb{R}^{N_MK}$, $\beta^x(t)$, $\beta^v(t)$, $t\in \mathbb{R}^+$}
    \State $\gamma^x \leftarrow \frac{\beta^x(t)}{1+\beta^x(t)}$
    \State $\boldsymbol{\mu}\sim\mathcal{N}(\gamma^x\frac{\hat{\mathbf{x}}_1+ \hat{\mathbf{x}}_2}{2}, \frac{1}{2}\gamma^x(1-\gamma^x\mathbf{I}))$
    \State $\mathbf{y}^v\sim\mathcal{N}(\frac{\beta^v(t)}{2}(K\mathbf{e}_{\hat{\mathbf{v}}_1}+K\mathbf{e}_{\hat{\mathbf{v}}_2}-2), \frac{\beta^v(t)}{2}K\mathbf{I})$
    \State $\boldsymbol{\theta}^v\leftarrow[\text{softmax}((\mathbf{y}^v)^{(d)})]_{d=1\dots N_M}$
    \State return $\boldsymbol{\mu}$, $\boldsymbol{\theta}^v$
    \EndFunction
    \Require Neural network $\boldsymbol{\Psi}$, $\mathcal{T}\mathbf{p}_1, \mathbf{p}_2 \in \mathbb{R}^{N_P(3+N_f)}$, $N, n_M, K\in\mathbb{N}^+, \sigma^x_1, \beta^v_1 \in \mathbb{R}^+$
    \State $\boldsymbol{\mu}\leftarrow\boldsymbol{0}$, $\rho \leftarrow1$, $\boldsymbol{\theta}^v\leftarrow[\frac{1}{K}]_{n_M\times K}$
    \For{$i=1$ to $N$}
    \State $t\leftarrow\frac{i-1}{n}$
    \State $\hat{\mathbf{x}}_1, \hat{\mathbf{v}}_1 \leftarrow p_O(\boldsymbol{\mu}, \boldsymbol{\theta}^v, \mathcal{T}\mathbf{p}_1, t;\boldsymbol{\Psi})$
    \State $\hat{\mathbf{x}}_2, \hat{\mathbf{v}}_2 \leftarrow p_O(\boldsymbol{\mu}, \boldsymbol{\theta}^v, \mathbf{p}_2, t;\boldsymbol{\Psi})$
    \State $\boldsymbol{\mu}, \boldsymbol{\theta}^v \leftarrow \text{update}(\hat{\mathbf{x}}_1, \hat{\mathbf{v}}_1, \hat{\mathbf{x}}_2, \hat{\mathbf{v}}_2, \sigma^x_1, \beta^v_1, t)$
    \EndFor
    \State $\hat{\mathbf{x}}_1, p_O^v(\hat{\mathbf{v}}_1|\mathbf{\boldsymbol\theta}^v, \mathbf{p}_1;1) \leftarrow p_O(\boldsymbol{\mu}, \boldsymbol{\theta}^v, \mathcal{T}\mathbf{p}_1, 1;\boldsymbol{\Psi})$
    \State $\hat{\mathbf{x}}_2, p_O^v(\hat{\mathbf{v}}_2|\mathbf{\boldsymbol\theta}^v, \mathbf{p}_2;1) \leftarrow p_O(\boldsymbol{\mu}, \boldsymbol{\theta}^v, \mathbf{p}_2, 1;\boldsymbol{\Psi})$
    \State $\hat{\mathbf{x}} \leftarrow \frac{\hat{\mathbf{x}}_1+\hat{\mathbf{x}}_2}{2}$
    \State $\hat{\mathbf{v}}\sim (p_O^v(\hat{\mathbf{v}}| \boldsymbol{\theta}^v, \mathcal{T}\mathbf{p}_1;1)+p_O^v(\hat{\mathbf{v}}| \boldsymbol{\theta}^v, \mathbf{p}_2;1))/2$
    \State return $[\hat{\mathbf{x}}, \hat{\mathbf{v}}]$
    \end{algorithmic}
\end{algorithm}

\section{Proofs}
\label{Proofs}
\subsection{Proof of Proposition 1}
\label{Proof of Proposition 1}

\textbf{Proposition 1.} \textit{In the fused Bayesian flow, the sender accuracies for continuous data are additive, i.e.,}
$\mathop{\mathbb{E}}\limits_{p_U(\boldsymbol{\theta}^x_{i-1}|\boldsymbol{\theta}^x_{i-2}, \cdot;\alpha^x_{i-1})}p_U(\boldsymbol{\theta}^x_i|\boldsymbol{\theta}^x_{i-1},\cdot;\alpha^x_i)=p_U(\boldsymbol{\theta}^x_i|\boldsymbol{\theta}^x_{i-2}, \cdot;\alpha^x_i+\alpha^x_{i-1}).$

\textit{Proof.} 
According to Eq. \ref{p_U of x}, given the original continuous data $\mathbf{x}_1$ and $\mathbf{x}_2$, the fused parameter $\boldsymbol{\mu}_i$ takes the following form:
\begin{equation}
\boldsymbol{\mu}_i \sim \mathcal{N}(\boldsymbol{\mu}_i|\frac{\alpha^x_i(\mathbf{x}_1+\mathbf{x}_2)}{2\rho_i}+\frac{\boldsymbol{\mu}_{i-1}\rho_{i-1}}{\rho_i},\frac{\alpha^x_i}{2\rho_i^2}\mathbf{I} ).
\end{equation}
Let 
\begin{equation}\boldsymbol{\mu}_i'\overset{\text{def}}{=} \frac{\alpha^x_i}{2\rho_i}(\mathbf{x}_1+\mathbf{x} _2)+\frac{\boldsymbol{\mu}_{i-1}\rho_{i-1}}{\rho_i},
\end{equation} 
then, using the reparameterization of the Gaussian distribution, $\boldsymbol{\mu}_i$ can be expressed as:
\begin{align}
\boldsymbol{\mu}_i = \boldsymbol{\mu}_i'+ \epsilon, \quad  \epsilon &\sim \mathcal{N}(0, \frac{\alpha^x_i}{2\rho_i^2}\mathbf{I}).
\end{align}
Since
\begin{equation}
    \boldsymbol{\mu}_{i-1} \sim \mathcal{N}(\boldsymbol{\mu}_i|\frac{\alpha^x_{i-1}}{2\rho_{i-1}}(\mathbf{x}_1+\mathbf{x} _2)+\frac{\boldsymbol{\mu}_{i-2}\rho_{i-2}}{\rho_{i-1}},\frac{\alpha^x_{i-1}}{2\rho_{i-1}^2}\mathbf{I}), \label{mu_{i-1}}
\end{equation}
replacing $\boldsymbol{\mu}_{i-1}$ in $\boldsymbol{\mu}'_i$ with Eq.\ref{mu_{i-1}} yields:
\begin{equation}
    \boldsymbol{\mu}'_i \sim \mathcal{N}(\boldsymbol{\mu}_i|\frac{\alpha^x_i+\alpha^x_{i-1}}{2\rho_i}(\mathbf{x}_1+\mathbf{x} _2)+\frac{\boldsymbol{\mu}_{i-2}\rho_{i-2}}{\rho_i},\frac{\alpha^x_{i-1}}{2\rho_i^2}\mathbf{I}).
\end{equation}
Therefore, we obtain 
\begin{equation}
    \boldsymbol{\mu}_i = \boldsymbol{\mu}_i'+ \epsilon \sim \mathcal{N}(\boldsymbol{\mu}_i|\frac{\alpha^x_i+\alpha^x_{i-1}}{2\rho_i}(\mathbf{x}_1+\mathbf{x} _2)+\frac{\boldsymbol{\mu}_{i-2}\rho_{i-2}}{\rho_i},\frac{\alpha^x_{i}+\alpha^x_{i-1}}{2\rho_i^2}\mathbf{I}),
\end{equation}
which implies
\begin{align}
p_U(\boldsymbol{\theta}^x_i|\boldsymbol{\theta}^x_{i-2}, \mathbf{x}_1, \mathbf{x}_2, \mathcal{T}\mathbf{p}_1, \mathbf{p}_2;\alpha^x_i+\alpha^x_{i-1}) &= \mathcal{N}(\boldsymbol{\mu}_i|\frac{\alpha^x_i+\alpha^x_{i-1}}{2\rho_i}(\mathbf{x}_1+\mathbf{x} _2)+\frac{\boldsymbol{\mu}_{i-2}\rho_{i-2}}{\rho_i},\frac{\alpha^x_{i}+\alpha^x_{i-1}}{2\rho_i^2}\mathbf{I}) \notag \\
&=\mathop{\mathbb{E}}\limits_{p_U(\boldsymbol{\theta}_{i-1}|\boldsymbol{\theta}^x_{i-2}, \mathbf{x}_1, \mathbf{x}_2, \mathcal{T}\mathbf{p}_1, \mathbf{p}_2;\alpha^x_{i-1})}p_U(\boldsymbol{\theta}^x_i|\boldsymbol{\theta}^x_{i-1},\mathbf{x}_1, \mathbf{x}_2, \mathcal{T}\mathbf{p}_1, \mathbf{p}_2;\alpha^x_i). \notag
\end{align}

\subsection{Proof of Proposition 2}
\label{Proof of Proposition 2}

\textbf{Proposition 2.} \textit{For discrete variables, the fused Bayesian flow satisfies the same additive property of accuracies, i.e.,} $\mathop{\mathbb{E}}\limits_{p_U(\boldsymbol{\theta}^v_{i-1}|\boldsymbol{\theta}^v_{i-2}, \cdot;\alpha^v_{i-1})}p_U(\boldsymbol{\theta}^v_i|\boldsymbol{\theta}^v_{i-1},\cdot;\alpha^v_i)=p_U(\boldsymbol{\theta}_i|\boldsymbol{\theta}^v_{i-2}, \cdot;\alpha^v_i+\alpha^v_{i-1}).$

\textit{Proof.} 
First, recall that
$\boldsymbol{\theta}^v_i = h(\mathbf{y}_i, \boldsymbol{\theta}^v_{i-1}), \mathbf{y}_i\sim p_S(\cdot|\mathbf{v}_1, \mathbf{v}_2;\alpha^v_i)$
and
$\boldsymbol{\theta}^v_{i-1} = h(\mathbf{y}_{i-1}, \boldsymbol{\theta}^v_{i-2}), \mathbf{y}_{i-1}\sim p_S(\cdot|\mathbf{v}^v_1, \mathbf{v}^v_2;\alpha^v_{i-1}).$
It then follows that
\begin{align}
\boldsymbol{\theta}^v_i= h(\mathbf{y}_i, \boldsymbol{\theta}^v_{i-1}
=h(\mathbf{y}_{i-1}, \boldsymbol{\theta}^v_{i-2}))&=\frac{\exp(\mathbf{y}_i)\frac{\exp(\mathbf{y}_{i-1})\boldsymbol{\theta}^v_{i-2}}{\sum^K_{k=1}\exp((\mathbf{y}_{i-1})_{k'})(\boldsymbol{\theta}^v_{i-2})_{k'}}}{\sum_{k=1}^K\exp((\mathbf{y}_i)_k)\frac{\exp((\mathbf{y_{i-1}})_k)(\boldsymbol{\theta}^v_{i-1})_k}{\sum_{k'=1}^K \exp((\mathbf{y_{i-1}})_{k'})(\boldsymbol{\theta}^v_{i-2})_{k'}}} \notag \\
&=\frac{\exp(\mathbf{y}_i+\mathbf{y}_{i-1})\boldsymbol{\theta}^v_{i-2}}{\sum^K_{k=1}\exp((\mathbf{y}_i+\mathbf{y}_{i-1})_k)(\boldsymbol\theta^v_{i-2})_k} \notag\\
&=h(\mathbf{y}_i+\mathbf{y}_{i-1}, \boldsymbol{\theta}^v_{i-2}).\label{h of i and i-1}
\end{align}
From Eq. \ref{fused p_s of v}, we have
\begin{align}
    \mathbf{y}_i \sim \mathcal{N}(\frac{\alpha^v_i}{2}(K\mathbf{e_{\mathbf{v}_1}}+K\mathbf{e_{\mathbf{v}_2}}-2), \frac{1}{2}\alpha^v_i K\mathbf{I}), \quad \mathbf{y}_{i-1} \sim \mathcal{N}(\frac{\alpha^v_{i-1}}{2}(K\mathbf{e_{\mathbf{v}_1}}+K\mathbf{e_{\mathbf{v}_2}}-2), \frac{1}{2}\alpha^v_{i-1} K\mathbf{I}). \notag
\end{align}
Therefore,
\begin{equation}
    \mathbf{y}_i+\mathbf{y}_{i-1}\sim \mathcal{N}(\frac{\alpha^v_i+\alpha^v_{i-1}}{2}(K\mathbf{e_{\mathbf{v}_1}}+K\mathbf{e_{\mathbf{v}_2}}-2), \frac{\alpha^v_i + \alpha^v_{i-1}}{2} K\mathbf{I}).
\end{equation}
Accordingly, the Bayesian update distribution associated with Eq. \ref{h of i and i-1} is
\begin{equation}
p_U(\boldsymbol{\theta}^v_i|\boldsymbol{\theta}^v_{i-1}, \mathbf{v}_1, \mathbf{v}_2, \mathcal{T}\mathbf{p}_1, \mathbf{p}_2;\alpha^v_i+\alpha^v_{i-1})=\mathop{\mathbb{E}}\limits_{\mathcal{N}(\frac{\alpha^v_i+\alpha^v_{i-1}}{2}(K\mathbf{e_{\mathbf{v}_1}}+K\mathbf{e_{\mathbf{v}_2}}-2), \frac{1}{2}(\alpha^v_i + \alpha^v_{i-1}) K\mathbf{I})} [\delta(\boldsymbol{\theta}^v_i-\frac{\mathbf{e}^{\mathbf{y}} \boldsymbol{\theta}^v_{i-2}}{ {\textstyle \sum_{k=1}^{K}}\mathbf{e} ^{\mathbf{y}_k}(\boldsymbol{\theta}^v_{i-2})_k })] \label{p_u of i and i-1}.
\end{equation}
Eq. \ref{p_u of i and i-1} accounts for all possible values of $\boldsymbol{\theta}^v_{i-1}$, that is, it marginalizes over $\boldsymbol{\theta}^v_{i-1}$. Therefore, it is in essence a marginal distribution, namely,
\begin{equation}
\mathop{\mathbb{E}}\limits_{p_U(\boldsymbol\theta^v_{i-1}|\boldsymbol{\theta}^v_{i-2}, \mathbf{v}_1, \mathbf{v}_2, \mathcal{T}\mathbf{p}_1, \mathbf{p}_2;\alpha^v_{i-1})}p_U(\boldsymbol\theta^v_i|\boldsymbol{\theta}^v_{i-1}, \mathbf{v}_1, \mathbf{v}_2, \mathcal{T}\mathbf{p}_1, \mathbf{p}_2;\alpha^v_i)) = p_U(\boldsymbol{\theta}^v_i|\boldsymbol{\theta}^v_{i-2}, \mathbf{v}_1, \mathbf{v}_2, \mathcal{T}\mathbf{p}_1, \mathbf{p}_2;\alpha^v_i+\alpha^v_{i-2}).\notag
\end{equation}

\subsection{Proof of Proposition 3}
\label{Proof of Proposition 3}

\textbf{Proposition 3.} \textit{The generative process preserves SE(3)-equivariant if $\boldsymbol{\Psi}$ is parameterized by an SE(3)-equivariant network and the Center of Mass (CoM) of the pair $(\mathcal{T}\mathbf{p}_1, \mathbf{p}_2)$ is shifted to zero.}

\textit{Proof.} Denote $T_g$ as the group of SE-(3) transformation, e.g. $T_g(\mathbf{x})=\mathbf{Rx}+\mathbf{b}$, where $\mathbf{R}  \in \mathbb{R} ^{3\times3}$ is the rotation matrix and $\mathbf{b} \in \mathbb{R} ^{3}$ is the translation vector. First, we move the pair $(\mathcal{T}\mathbf{p}_1, \mathbf{p}_2)$ as a whole to the position where the Center of Mass (CoM) is at the origin, applying the same translation to the relevant variables. At this point, translation equivariance is satisfied by definition, and it is only necessary to ensure rotational equivariance (i.e. O(3)-equivariance). Since atom types are inherently invariant under SE(3)-transformations, it suffices to consider only how the atom coordinates transform. For simplicity, the following discussion considers only variables in 3D space, ignoring those related to atom types and $t$.


Given $[\hat{\mathbf{x}}_1, \hat{\mathbf{x}}_2] = \boldsymbol{\Psi}(\boldsymbol{\theta}, \mathcal{T}\mathbf{p}_1, \mathbf{p}_2)$, when 
$T_g$ is applied to $\boldsymbol{\theta}, \mathcal{T}\mathbf{p}_1, \mathbf{p}_2$, since $\boldsymbol{\Psi}$ is an SE(3)-equivariant network, the output of the network will be:
\begin{align}
\boldsymbol{\Psi}(T_g(\boldsymbol{\theta}, \mathcal{T}\mathbf{p}_1, \mathbf{p}_2))&=\boldsymbol{\Psi}(\mathbf{R}(\boldsymbol{\theta}, \mathcal{T}\mathbf{p}_1, \mathbf{p}_2))=\mathbf{R\Psi}(\boldsymbol{\theta}, \mathcal{T}\mathbf{p}_1, \mathbf{p}_2)\notag \\
&=\mathbf{R}([\hat{\mathbf{x}}_1, \hat{\mathbf{x}}_2])=[\mathbf{R}\hat{\mathbf{x}}_1, \mathbf{R}\hat{\mathbf{x}}_2]=[T_g(\hat{\mathbf{x}}_1), T_g(\hat{\mathbf{x}}_2)].\label{eq 28}
\end{align}
From Eq. \ref{p_F of x}, the transition density of the parameters is given by:
\begin{equation}
p_F(\boldsymbol{\theta}_i|\hat{\mathbf{x}}_1, \hat{\mathbf{x}}_2, \mathcal{T}\mathbf{p}_1, \mathbf{p}_2) = \mathcal{N}(\boldsymbol{\mu}_i|\gamma\frac{\hat{\mathbf{x}}_1+ \hat{\mathbf{x}}_2}{2}, \frac{1}{2}\gamma(1-\gamma)\mathbf{I}).\notag
\end{equation}
We can then prove that it remains O(3)-invariant during the generative process:
\begin{align}
&p_F(T_g(\boldsymbol{\theta}_i)|T_g(\hat{\mathbf{x}}_1), T_g(\hat{\mathbf{x}}_2), T_g(\mathcal{T}\mathbf{p}_1), T_g(\mathbf{p}_2))\notag\\ 
   =& \mathcal{N}(T_g(\boldsymbol{\mu}_i)|\gamma\frac{T_g(\hat{\mathbf{x}}_1)+ T_g(\hat{\mathbf{x}}_2)}{2}, \frac{1}{2}\gamma(1-\gamma)\mathbf{I})\notag\\
   =& \mathcal{N}(\mathbf{R}(\boldsymbol{\mu}_i)|\gamma\frac{\mathbf{R}(\hat{\mathbf{x}}_1)+ \mathbf{R}(\hat{\mathbf{x}}_2)}{2}, \frac{1}{2}\gamma(1-\gamma)\mathbf{I}) \quad \text{(according to Eq.\ref{eq 28})} \notag\\
   =& \mathcal{N}(\boldsymbol{\mu}_i|\gamma\frac{\hat{\mathbf{x}}_1+ \hat{\mathbf{x}}_2}{2}, \frac{1}{2}\gamma(1-\gamma)\mathbf{I}) \quad
   \text{(equivariance of isotropic Gaussian)}\notag\\
=&p_F(\boldsymbol{\theta}_i|\hat{\mathbf{x}}_1, \hat{\mathbf{x}}_2, \mathcal{T}\mathbf{p}_1, \mathbf{p}_2).
\end{align}

\section{Implementation Details}
\label{Implementation Details}

\subsection{Featurization}

Each protein atom is described by a feature vector consisting of a one-hot encoding of the element type (H, C, N, O, S, Se), a 20-dimensional one-hot vector indicating the amino acid type, a one-dimensional indicator specifying whether the atom belongs to the protein backbone, and a one-hot arm/scaffold region indicator, which is determined by the atom’s distance to the center of the arm prior. Ligand atoms are represented by one-hot vectors over the element set $\{$C, N, O, F, P, S, Cl$\}$, together with aromatic information.

We adopt the single-target drug design method MolCRAFT \citep{qu2024molcraft} as the backbone in our framework. Following MolCRAFT, two graphs are dynamically constructed for message passing in the protein–ligand complex: a $k$-nearest-neighbors (knn) graph over ligand and protein atoms, and a fully connected graph over ligand atoms. $k$ is set to 32. In the knn graph, edge features are defined as the outer product of a distance embedding and an edge-type embedding, where the distance embedding is computed by expanding pairwise distances with radial basis functions, and the edge type is represented as a 4-dimensional one-hot vector. In the ligand graph, bond information is encoded by a one-hot vector with five categories: non-bond, single, double, triple, and aromatic.

\subsection{SE(3)-Equivariant Network}

The interactions between protein pocket atoms and molecule atoms are modeled by an SE(3)-equivariant graph neural network \citep{targetdiff}. At layer
$l$, the hidden features $\mathbf{h}^{l}$ and coordinates $\mathbf{x}^{l}$ are updated as
\begin{align} \mathbf{h}_i^{l+1}&=\mathbf{h}_i^{l}+\sum_{j\in\mathcal{V},i\neq j}f_h(d_{ij}^l, \mathbf{h}_i^{l},\mathbf{h}_j^{l},\mathbf{e}_{ij};\theta_h) \\ \mathbf{x}_i^{l+1}&=\mathbf{x}_i^{l}+\sum_{j\in\mathcal{V},i\neq j}(\mathbf{x}_i^l-\mathbf{x}_j^l)f_x(d_{ij}^l,\mathbf{h}_i^{l+1},\mathbf{h}_j^{l+1},\mathbf{e}_{ij};\theta_x)\cdot \mathbf{l}_{mask},\end{align}
where $l$ denotes the layer index, $\mathcal{V}$ denotes the set of all atoms in the protein–ligand complex, $d_{ij}=||\mathbf{x}_i-\mathbf{x}_j||$ and $\mathbf{e}_{ij}$ represents the relative distance information together with option edge features between atoms $i$ and $j$. The mask $\mathbf{l}_{mask}$ is applied to ligand atoms to remain the protein atom coordinates fixed during message passing. $f_h$ and $f_x$ are implemented as attention blocks.

For target $\mathcal{P}_1$, $\boldsymbol{\Psi}$ takes as input the initial node coordinates $\mathbf{x}^0 = [\boldsymbol{\mu}, \mathbf{x}_{P_1} ]$, and the initial node features $\mathbf{h}_0 = \text{linear}(\boldsymbol\theta^v, \mathbf{v}_{P_1} , t)$. At the final layer, $\boldsymbol{\Psi}$ outputs the coordinate estimate $\hat{\mathbf{x}} = \boldsymbol{\Psi}_{P_1}^x$. For the discrete variables, the predicted distribution is obtained by applying a softmax function to the corresponding network output $\boldsymbol{\mathbf{\hat{v}}}^{(d)} = \text{softmax} ((\boldsymbol{\Psi}_{P_1}^v)^{(d)})$. The same procedure is applied to target $\mathcal{P}_2$.

\subsection{Model Information}

We use an SE(3)-equivariant network with 9 equivariant layers, each implemented as a transformer with hidden dimension 128 and 16 attention heads. The key/value embeddings and attention scores are parameterized by 2-layer MLPs with ReLU activation and Layer Normalization. 

For the noise schedules, we set $\beta_1=1.5$ for atom types and $\sigma_1=0.03$ for atom coordinates. During generation, the number of sampling steps is set to 100. 

\subsection{Alignment}

To maintain consistency with previous work \citep{zhou2024reprogramming}, we use TargetDiff to generate the probe ligands. During probe-ligand alignment, the weights are set to 
$\lambda_{heavy}=1$ and $\lambda_{light}=0.1$. For direct alignment of protein pockets, RANSAC is performed for 1000 iterations, followed by ICP with a maximum of 50 iterations.


\section{More Experimental Results}
\subsection{Baselines on Single-Target Setting}

To compare the performance of TargetDiff and MolCRAFT on single-target drug design, we generate 10 molecules for each of the 438 unique targets and evaluate the resulting molecules in terms of binding affinity (Vina Score, Vina Min, Vina Dock, and High Affinity) and molecular properties (drug-likeness QED, synthesizability SA, and diversity). We use the widely adopted AutoDock Vina \citep{trott2010autodock} to estimate the mean, trimmed mean (i.e., averaging that removes the top $10\%$ and bottom $10\%$ of values before calculating) and median (denoted as “Avg.”, “T-Avg.” and “Med.” respectively) of affinity-related metrics. Vina Score evaluates binding affinity based on the generated poses directly; Vina Min first applies local minimization to the pose and then estimates the affinity; Vina Dock uses a re-docking procedure to approximate the optimal binding affinity; and High Affinity measures the percentage of generated molecules whose binding affinity is better than that of the reference ligand for each test protein.

As shown in Table \ref{single target}, MolCRAFT generates molecules with higher binding affinity and more favorable molecular properties. This advantage stems from the fact that the Bayesian Flow Network models continuous atomic coordinates and discrete atom types jointly in a unified continuous parameter space. By contrast, TargetDiff is less effective at handling such mixed-modality data, leading to weaker generation quality.

\begin{table}[htbp]
\vspace{-1em}
\begin{center}
\captionsetup{skip=4pt}
\caption{Summary of the binding affinity and molecular properties of reference ligands and molecules generated by baselines in the single-target setting. $(\uparrow) / (\downarrow)$ indicates that higher/lower values are preferred. Since the mean is highly sensitive to outliers, some abnormal values of Vina Score and Vina Min are excluded.}
\renewcommand{\arraystretch}{1.2}
\label{single target}
\resizebox{1\columnwidth}{!}{
\begin{tabular}{c|ccc|ccc|ccc|cc|cc|cc|cc}
\toprule
\multirow{2}{*}{Methods} & \multicolumn{3}{c|}{Vina Score $(\downarrow)$} & \multicolumn{3}{c|}{Vina Min $(\downarrow)$} & \multicolumn{3}{c|}{Vina Dock $(\downarrow)$} & \multicolumn{2}{c|}{High Affinity $(\uparrow)$} & \multicolumn{2}{c|}{QED $(\uparrow)$} & \multicolumn{2}{c|}{SA $(\uparrow)$} & \multicolumn{2}{c}{Diversity $(\uparrow)$} \\
                         & Avg.              & T-Avd.       & Med.        & Avg.             & T-Avd.       & Med.       & Avg.       & T-Avd.      & Med.               & Avg.                   & Med.                   & Avg.              & Med.              & Avg.              & Med.             & Avg.                 & Med.                \\ \midrule
Reference                      & -                 & -8.05        & -7.99       & -                 & -8.10        & -8.10      & -8.06      & -8.38       & -8.29              & -                      & -                      & 0.54              & 0.55              & 0.74              & 0.78             & -                    & -                   \\ \midrule
TargetDiff               & -                 & -7.38        & -7.38       & -                & -7.89        & -7.87      & -8.76      & -8.78       & -8.76              & 56.5\%                 & 60.0\%                 & 0.51              & 0.53              & 0.58              & 0.58             & 0.69                 & 0.70                \\
MolCRAFT                 & -        & -8.33        & -8.28       & -       & -8.53        & -8.47      & -9.20      & -9.11       & -9.03     & 65.0\%        & 70.0\%        & 0.52              & 0.53              & 0.66              & 0.65             & 0.69                 & 0.68                \\ \midrule
\end{tabular}
}
\end{center}
\vspace{-1em}
\end{table}

\subsection{Comparison of Prober Generators}

We compare different methods for generating the probe ligand used in prior-based pocket alignment. Specifically, we use TargetDiff and MolCRAFT, respectively, to generate probe ligands, and then align their docked poses on $\mathcal{P}_1$
 and $\mathcal{P}_2$
 to estimate the pocket transformation. We randomly sample 1,000 dual-target pairs from the dataset and generate 10 molecules for each pair. The results are reported in Table \ref{prober}. As can be seen, the quality of the generated probe ligand affects the alignment accuracy and consequently the dual-target generation performance. Using MolCRAFT to generate the prober leads to better overall results than using TargetDiff. This is because that the single-target molecules generated by MolCRAFT have higher quality, which can more effectively indicate the specific spatial alignment between the two protein pockets.


\begin{table}[htbp]
\begin{center}
\captionsetup{skip=4pt}
\caption{Comparison of probe ligands generated by different methods for prior-based pocket alignment.}
\renewcommand{\arraystretch}{1.2}
\label{prober}
\resizebox{1\columnwidth}{!}{
\begin{tabular}{c|cc|cc|cc|cc|cc|cc|cc}
\toprule
\multirow{2}{*}{Methods} & \multicolumn{2}{c|}{P-1 Vina Dock $(\downarrow)$} & \multicolumn{2}{c|}{P-2 Vina Dock $(\downarrow)$} & \multicolumn{2}{c|}{Max Vina Dock $(\downarrow)$} & \multicolumn{2}{c|}{Dual High Aff. $(\uparrow)$} & \multicolumn{2}{c|}{QED $(\uparrow)$} & \multicolumn{2}{c|}{SA $(\uparrow)$} & \multicolumn{2}{c}{Diversity $(\uparrow)$} \\
                         & Avg.                    & Med.                    & Avg.                    & Med.                    & Avg.                    & Med.                    & Avg.                    & Med.                   & Avg.              & Med.              & Avg.              & Med.             & Avg.                 & Med.                \\ \midrule
TargetDiff-prior         & -8.71                   & -8.72                   & -8.74                   & -8.74                   & -8.03                   & -8.12                   & 56.2\%                  & 60.0\%                 & 0.55              & 0.57              & 0.68              & 0.68             & 0.69                 & 0.69                \\
MolCRAFT-prior           & -8.85                   & -8.79                   & -8.85                   & -8.80                   & -8.18                   & -8.19                   & 58.1\%                  & 60.0\%                 & 0.55              & 0.57              & 0.68              & 0.68             & 0.69                 & 0.69                \\ \midrule
\end{tabular}
}
\end{center}
\end{table}

\subsection{Significance Test on Affinity Metrics}

\label{Significance Test on Affinity Metrics}

To further verify the superior affinity performance of FusedBFN, we conduct paired t-tests between FusedBFN and the second-best baseline, DualDiff, on the affinity-related metrics, where the results of the two methods are compared for each target pair. Note that lower values are preferred for P-1 Vina Dock, P-2 Vina Dock, and Max Vina Dock, whereas higher values are better for Dual High Affinity. As shown in Table \ref{t-test}, all tested metrics yield p-values far below 0.05, indicating that FusedBFN achieves statistically significant improvements over DualDiff in terms of dual-target binding affinity.

\begin{table}[htbp]
\vspace{-1em}
\begin{center}
\captionsetup{skip=4pt}
\caption{Paired t-test results between FusedBFN and DualDiff on dual-target affinity-related metrics.}
\renewcommand{\arraystretch}{1.2}
\label{t-test}
\begin{tabular}{c|c|c}
\toprule
Metric & $t$ & $p$ \\ \midrule
P-1 Vina Dock & -20.81  & $6.52 \times 10^{-95}$ \\
P-2 Vina Dock & -27.56  & $ 7.69 \times 10^{-163}$ \\
Max Vina Dock & -23.88  & $1.16 \times 10^{-123}$ \\
Dual High Aff. & 36.22   & $6.33 \times 10^{-274}$ \\ \midrule
\end{tabular}
\end{center}
\vspace{-1em}
\end{table}

\subsection{Sampling Efficiency}
\label{Sampling Efficiency}

We further compare the sampling efficiency of different methods in Table \ref{sample time}, which reports the average time required to generate 10 molecules for one dual-target pair on a TITAN Xp GPU with batch size set to 10. FusedBFN takes 113.68 seconds on average, which is substantially faster than DualDiff (1111.28 seconds), achieving nearly a $10\times$ speedup. Compared with the single-target method MolCRAFT, FusedBFN takes about twice as much time. It is consistent with our design since FusedBFN constructs two protein-ligand graphs for dual-target generation. In addition, the message passing on the two graphs can be executed in parallel when GPU memory is sufficient, offering additional room for acceleration during inference.

\begin{table}[htbp]
\vspace{-1em}
\begin{center}
\captionsetup{skip=4pt}
\caption{Inference time required by different methods to generate 10 molecules. Lower values indicate higher generation efficiency.}
\renewcommand{\arraystretch}{1.2}
\label{sample time}
\begin{tabular}{l|l|l|l}
\toprule
     Method & MolCRAFT                   & DualDiff                     & FusedBFN                   \\ \midrule
Time (s) & \multicolumn{1}{c|}{57.27} & \multicolumn{1}{c|}{1128.40} & \multicolumn{1}{c}{113.68} \\ \midrule
\end{tabular}
\end{center}
\vspace{-1em}
\end{table}

\section{Limitations, Future Work and Broader Impact}

\label{Limitations and Future Work}

Despite the outstanding performance of FusedBFN, several limitations remain. First, FusedBFN depends on the quality of dual-target alignment, and inappropriate alignment may negatively affect the dual-target information fusion. Second, our framework extends a pretrained single-target model to the dual-target setting, which may limit its capacity to fully capture the binding patterns between molecules and dual-target simultaneously.  In future work, we plan to construct larger structural datasets for dual-target complexes and train on such data directly. We are also interested in extending the current framework to more general multi-target molecular design scenarios.

Our work holds promise for accelerating the discovery of polypharmacological compounds for complex diseases. However, the generated molecules are still computational hypotheses and should be used with caution, as rigorous downstream validation is required for synthesis feasibility, safety, and off-target effects.

\section{More Visualization Examples}
\label{More Visualization Examples}

\begin{figure}[htbp]
  \centering
\includegraphics[width=1\linewidth]{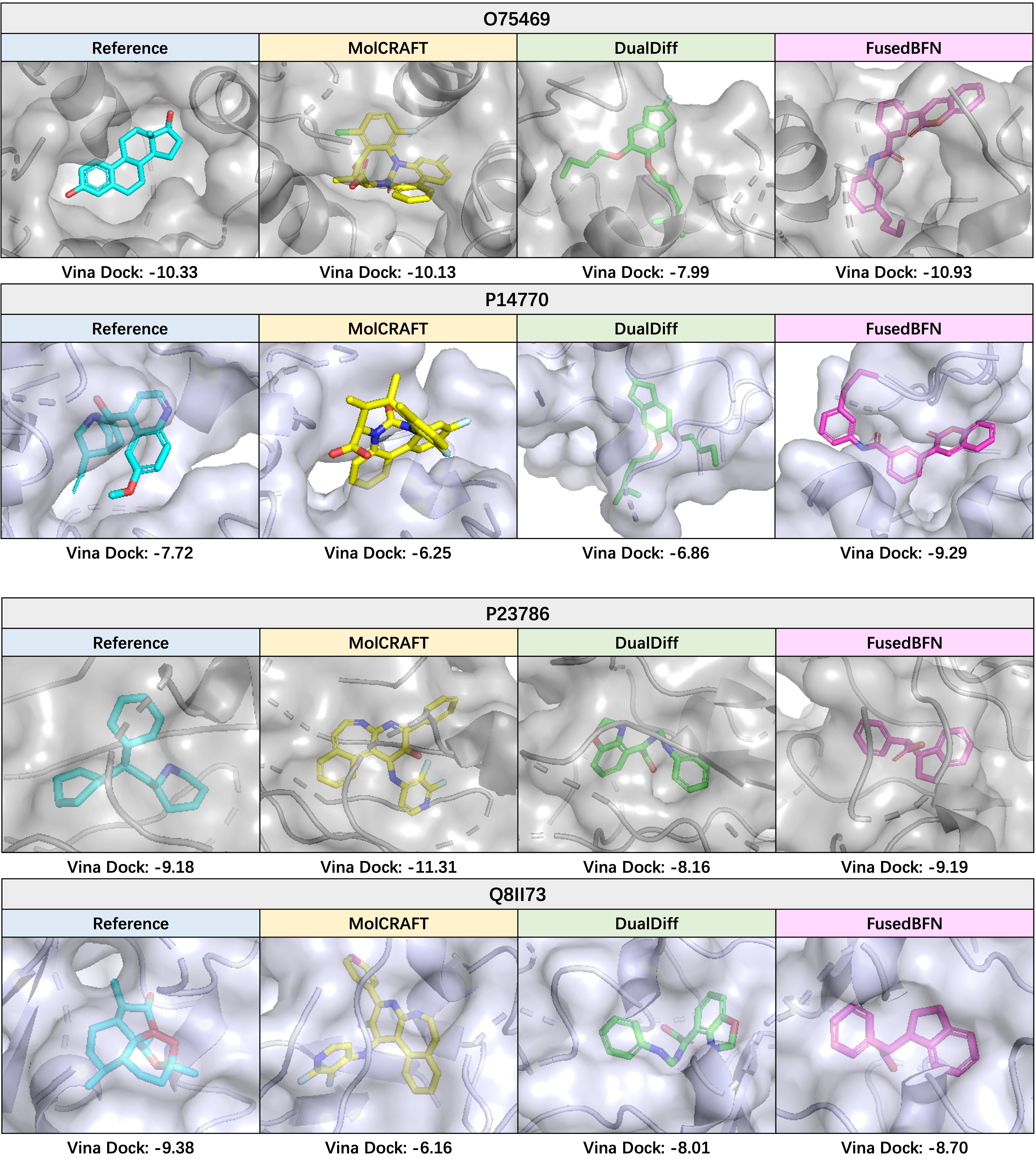}
  \caption{Visualization of more reference ligands and molecules generated by MolCRAFT, DualDiff and FusedBFN.}
  \label{visualization_appendix}
\end{figure}

\end{document}